\documentclass[journal]{IEEEtran}
\usepackage{amsmath,amsfonts}
\usepackage{hyperref}
\usepackage{algorithm}
\usepackage{algpseudocode} 
\usepackage{array}
\usepackage[caption=false,font=normalsize,labelfont=sf,textfont=sf]{subfig}
\usepackage{textcomp}
\usepackage{stfloats}
\usepackage{url}
\usepackage{verbatim}
\usepackage{graphicx}
\usepackage{cite}
\usepackage{amsthm}
\usepackage{xcolor}
\usepackage{marginnote}
\usepackage[normalem]{ulem}
\usepackage{multirow}

\newtheorem{thm}{Theorem}[section]
\newtheorem{prop}[thm]{Proposition}
\newtheorem{lem}[thm]{Lemma}

\begin{document}

\title{Memory-efficient model-based deep learning with convergence and robustness guarantees}
\author{Aniket Pramanik,~\IEEEmembership{Student Member,~IEEE,} M. Bridget Zimmerman, Mathews Jacob,~\IEEEmembership{Fellow,~IEEE}
\thanks{Aniket Pramanik and Mathews Jacob are from the Department of Electrical and Computer Engineering at the University of Iowa, Iowa City, IA, 52242, USA (e-mail: aniket-pramanik@uiowa.edu; mathews-jacob@uiowa.edu). M. Bridget Zimmerman is from the Department of Biostatistics at the University of Iowa, Iowa City, IA, 52242, USA (e-mail: bridget-zimmerman@uiowa.edu). This work is supported by grants NIH R01 AG067078 and R01 EB031169.}}

\markboth{Journal of \LaTeX\ Class Files,~Vol.~14, No.~8, August~2021}%
{Shell \MakeLowercase{\textit{et al.}}: A Sample Article Using IEEEtran.cls for IEEE Journals}


\maketitle

\begin{abstract}
Computational imaging has been revolutionized by compressed sensing algorithms, which offer guaranteed uniqueness, convergence, and stability properties. Model-based deep learning methods that combine imaging physics with learned regularization priors have emerged as more powerful alternatives for image recovery. The main focus of this paper is to introduce a memory efficient model-based algorithm with similar theoretical guarantees as CS methods. The proposed iterative algorithm alternates between a gradient descent involving the score function and a conjugate gradient algorithm to encourage data consistency. The score function is modeled as a monotone convolutional neural network. Our analysis shows that the monotone constraint is necessary and sufficient to enforce the uniqueness of the fixed point in arbitrary inverse problems. In addition, it also guarantees the convergence to a fixed point, which is robust to input perturbations. 
We introduce two implementations of the proposed MOL framework, which differ in the way the monotone property is imposed. The first approach enforces a strict monotone constraint, while the second one relies on an approximation. The guarantees are not valid for the second approach in the strict sense. However, our empirical studies show that the convergence and robustness of both approaches are comparable, while the less constrained approximate implementation offers better performance. The proposed deep equilibrium formulation is significantly more memory efficient than unrolled methods, which allows us to apply it to 3D or 2D+time problems that current unrolled algorithms cannot handle. \end{abstract}


\section{Introduction}
\noindent The recovery of images from a few noisy measurements is a common problem in several imaging modalities, including MRI \cite{fessler2010model}, CT \cite{elbakri2002statistical}, PET \cite{verhaeghe2008dynamic}, and microscopy \cite{aguet2008model}. In the undersampled setting, multiple images can give a similar fit to the measured data, making the recovery ill-posed. Compressive sensing (CS) algorithms pose the recovery as a convex optimization problem, where a strongly convex prior is added to the data-consistency term to regularize the recovery \cite{lustig2007sparse}. The main benefit of convex priors is in the uniqueness of the solutions. In particular, the strong convexity of the priors guarantees that the overall cost function in \eqref{eqn:cs} is strongly convex, even when $\mathbf A$ operator has a large null space. Another desirable property of convex priors in CS is the robustness of the solution to input perturbations. 

In recent years, several flavors of model-based deep learning algorithms, which combine imaging physics with learned priors, were introduced to significantly improve the performance compared to CS algorithms. For example, plug and play (PnP) methods use denoiser modules to replace the proximal mapping steps in CS algorithms \cite{venkatakrishnan2013plug,buzzard2018plug,romano2017little,sun2019online,ryu2019plug,sun2021scalable}, and the algorithms are run until convergence. While earlier approaches chose off-the-shelf denoisers such as BM3D \cite{dabov2009bm3d}, recent methods use  pre-trained  convolutional neural network (CNN) modules \cite{sun2019online,ryu2019plug}. The pre-trained CNN modules that learn the image prior are agnostic to the forward model, which enables their use in arbitrary inverse problems. These methods come with convergence and uniqueness guarantees when the forward model is full-rank or the data term is strongly convex \cite{kamilov2023plug}. When the data term is not strongly convex, weaker convergence guarantees are available \cite{sun2019online}, but uniqueness is not guaranteed. 
Another category of approaches relies on unrolled optimization; these algorithms unroll finite number of iterative optimization steps in CS algorithms to obtain a deep network, which is composed of CNN blocks and optimization blocks to enforce data consistency; the resulting deep network is trained in an end-to-end fashion \cite{gregor2010learning,hammernik2018learning,aggarwal2018modl, xiang2021fista}. A key difference between unrolled methods and PnP methods is that the CNN block is trained end-to-end, assuming a specific forward model; such model-based methods typically offer better performance than PnP methods that are agnostic to the forward model \cite{gregor2010learning,hammernik2018learning,schlemper2017deep, aggarwal2018modl, sun2016deep, xiang2021fista,ongie2020deep}. Unlike PnP approaches that run the algorithm until convergence, the number of iterations in unrolled methods are restricted by the memory of the GPU devices during training; this often limits the applicability of unrolled algorithms to large-scale multi-dimensional problems. Several strategies were introduced to overcome the memory limitations of unrolled methods.  For an unrolled network with $N$ iterations and shared CNN modules across iterations, the computational complexity and memory demand of backpropagation are $\mathcal O(N)$ and $\mathcal O(N)$, respectively. The forward steps can be recomputed during backpropagation, which reduces the memory demand to $\mathcal O(1)$, while the computational complexity increases to $\mathcal O(N^2)$. Forward checkpointing \cite{chen2016training} saves the variables for every $K$ layers during forward propagation, which reduces the computational demand to $\mathcal O(NK)$, while the memory demand is $\mathcal O(N/K)$. Reverse recalculation has been proposed to reduce the memory demand to $\mathcal O(1)$ and computational complexity to $\mathcal O(N)$ \cite{kellman2020memory}. However, the approach in \cite{kellman2020memory} requires multiple iterations to invert each CNN block, resulting in high computational complexity in practical applications.

 Gilton et al. recently extended the deep equilibrium (DEQ) model \cite{bai2019deep} to significantly improve the memory demand \cite{gilton2021deep} of unrolled methods. Unlike unrolled methods, DEQ schemes run the iterations until convergence, similar to PnP algorithms. This property allows one to perform forward and backward propagation using fixed-point iteration involving a single physical layer, which reduces the memory demand to $\mathcal O(1)$, while the computational complexity is $\mathcal O(N)$; this offers better tradeoffs than the alternatives discussed above \cite{chen2016training,kellman2020memory}. The runtime of DEQ methods that are iterated until convergence are variable compared to unrolled methods, which use a finite number of iterations. In addition, the convergence of the iterative algorithm is crucial for the accuracy of backpropagation steps in DEQ, unlike in unrolled methods. Convergence guarantees were introduced in  \cite{ryu2019plug,gilton2021deep} for the alternating direction method of multipliers (ADMM),  proximal gradient (PG), and forward-backward DEQ algorithms.  The convergence guarantees rely on restrictive conditions on the CNN denoising blocks, which are dependent on the forward models. Unfortunately, when the minimum singular value of the forward operator is small (e.g., highly accelerated parallel MRI) or zero (e.g., super-resolution), the CNN denoiser needs to be close to an identity operator for the iterations to converge. Another challenge associated with DEQ methods is the way the non-expansive constraints on the network are imposed.  Most methods \cite{ryu2019plug,gilton2021deep}  use spectral normalization of each layer of the network. Our experiments in Fig. \ref{fig:brain_train_plots} show that spectral normalization often translates to networks with lower performance. Another theoretical problem associated with current DEQ methods is the potential non-uniqueness of the fixed point, which can also affect the stability/robustness of the algorithm in the presence of input perturbations. 
 We note that the stability of deep image reconstruction networks is a debated topic. While  deep networks are reported to be more fragile to input perturbations than are conventional algorithms \cite{antun2020instabilities}, some of the recent works have presented a more optimistic view \cite{darestani2021measuring, genzel2022solving}. 

The main goal of this work is to introduce a model-based DEQ algorithm that shares the desirable properties of convex CS algorithms, including guaranteed uniqueness of the fixed point solutions, convergence, and robustness to input perturbations. By enabling the training of the CNN modules in an end-to-end fashion, the proposed algorithm can match the performance of unrolled approaches while being significantly more memory efficient. The main contributions of this paper are:
\begin{itemize}
	\item We introduce a forward-backward DEQ algorithm \eqref{fp} involving a learned network $\mathcal F$. Existing algorithms \cite{romano2017little,aggarwal2018modl,hammernik2020machine} such as MoDL and RED are special cases of this algorithm when the damping parameter $\alpha=1$. 
	
	\item We show that constraining the CNN module as an $m>0$ monotone operator is necessary and sufficient to guarantee the uniqueness of the fixed point of the algorithm. Because the monotone constraint is central to our approach, we term the proposed scheme as the monotone operator learning (MOL) algorithm.
	
	\item We show that an $m$-monotone operator $\mathcal F$ can be realized as a residual CNN: $\mathcal F=\mathcal I-\mathcal H_{\theta}$, where the Lipschitz constant of the denoiser module $\mathcal H_{\theta}$ is $L[\mathcal H_{\theta}]=1-m$. We also determine the range of values of $\alpha$ and $L[\mathcal H_{\theta}]$ for which the algorithm converges; the analysis and the experiments in Fig. \ref{fig:brain_train_plots} show that the direct application of the MoDL and RED ($\alpha=1$) algorithms to the DEQ setting will diverge unless a highly constrained CNN ($L[\mathcal H_{\theta}]<0.24$) is used, which restricts performance. By contrast, the use of a smaller $\alpha$ translates to higher $L[\mathcal H_{\theta}]$ and hence improved performance.
	\item We theoretically analyze the worst-case sensitivity of the the resulting DEQ scheme. Our analysis shows that the norm of the perturbations in the reconstructed images are linearly proportional to the norm of the measurement perturbations, with the proportionality dependent on $1/m$. 
\item  We introduce two implementations of the proposed MOL algorithm. The first approach uses spectral normalization to enforce the monotone constraint in the strict sense. We also introduce an approximate implementation, where we replace $L[\mathcal H_{\theta}]$ by an approximation $l[\mathcal H_{\theta}]$. While the second approach does not satisfy the monotone constraint in the strict sense, our experiments in Fig. \ref{fig:brain_train_plots} shows that the resulting algorithm converges, while Fig. \ref{fig:brain_rob_compare} shows that the robustness of both schemes to adversarial and Gaussian noise are comparable. We note that spectral normalization based estimate for Lipschitz constant is very conservative; our experiments in Fig. \ref{fig:knee_perf_compare} show that the second approach offers improved performance over the exact approach.
	\item We experimentally compare the performance against unrolled algorithms that use similar-sized CNNs in two-dimensional MR imaging problems. Our results show that the performance of the MOL scheme is comparable to that of unrolled algorithms. In addition, the MOL scheme is associated with a ten-fold reduction in memory compared to the unrolling algorithms with ten unrolls. The significant gain in memory demand allows us to extend our algorithm to the 3D or 2D+time setting, where it offers improved performance over unrolled 2D approaches. The experimental results in Fig. \ref{fig:brain_rob_compare} and \ref{fig:brain_robustness_plots} show the improved robustness of the proposed scheme compared to existing unrolled algorithms \cite{aggarwal2018modl,sun2016deep} and UNET \cite{ronneberger2015u}. The recorded run-times in Table \ref{tab:perf_comp_tab} show that MOL has higher computational complexity ($\approx$ 2.5 times) compared to unrolling algorithms due to more iterations, when compared with fixed number of unrolls in the latter. Our experiments in Fig. \ref{fig:brain_rob_compare} show that the increased computational complexity translate to an improvement in robustness performance over unrolled algorithms, when Lipschitz regularization is applied on the networks. 
\end{itemize}

\section{Background}
We consider recovery of an image $\mathbf x$ from its noisy undersampled measurements $\mathbf b$, specified by
\begin{equation}
\label{eq:linear}
\mathbf b = \mathbf A\mathbf x+ \mathbf n,
\end{equation}
where $\mathbf A$ is a linear operator and $\mathbf n \sim \mathcal N(\mathbf 0, \sigma^2 \mathbf I)$ is an additive white Gaussian noise. The measurement model provides a conditional probability $p(\mathbf b|\mathbf x) = \mathcal N(\mathbf A\mathbf x, \sigma^2 \mathbf I)$. The maximum a posteriori (MAP) estimation of $\mathbf x$ from the measurements $\mathbf b$ poses the recovery as 
\begin{equation}
\label{eq:map}
    \mathbf x_{\rm MAP} = \arg \max_{\mathbf x} \log p(\mathbf x|\mathbf b).
\end{equation}
Using Bayes' rule, $p(\mathbf x|\mathbf b) \propto p(\mathbf b|\mathbf x)p(\mathbf x)$ and from the monotonicity of the $\log$ function, one obtains $$\mathbf x_{\rm MAP}= \arg \min_{\mathbf x} C(\mathbf x),$$ where
\begin{equation}
\label{eqn:cs}
C(\mathbf x) = \underbrace{\frac{\lambda}{2}\|\mathbf A\mathbf x - \mathbf b\|_2^2}_{\mathcal D(\mathbf x) = -\log~p(\mathbf b|\mathbf x)} + \underbrace{\phi(\mathbf x)}_{-\log p(\mathbf x)}. 
\end{equation}
Here, $\lambda = \frac{1}{\sigma^2}$. The first term $\mathcal D(\mathbf x) = -\log~p(\mathbf b|\mathbf x)$ is the data-consistency term, while the second term is the log prior. Compressed sensing algorithms use convex prior distributions (e.g., $\phi(\mathbf x) = \|\mathbf x\|_{\ell_1}$) to result in a strongly convex cost function with unique minimum. 

We note that the minimum of \eqref{eqn:cs} satisfies the fixed-point relation:
\begin{equation}
\label{eqn:grad}
\underbrace{\lambda \mathbf A^H (\mathbf A\mathbf x - \mathbf b)}_{\mathcal G(\mathbf x)} + \underbrace{\nabla_{\mathbf x}~\phi(\mathbf x)}_{\mathcal F(\mathbf x)} = 0,
\end{equation}
where $\mathbf A^H$ is the Hermitian operator of $\mathbf A$. We note that the first term $\mathcal G(\mathbf x)$ is the noise in the measurements, translated to the image domain. When the above fixed-point relation holds, $\mathcal F(\mathbf x)$ is essentially a noise estimator, often referred to as the score function. $\mathcal F$ points to the maximum of the prior $p(\mathbf x)$. 

Several algorithms that converge to the fixed point of \eqref{eqn:grad} have been introduced in the CS setting \cite{yang2010fast,daubechies2004iterative,beck2009fast}. For example, forward-backward algorithms rewrite \eqref{eqn:grad} as $
(\mathcal I + \alpha \mathcal F)(\mathbf x) = (\mathcal I - \alpha \mathcal G)(\mathbf x), ~ \alpha > 0$, which has the same fixed point as \eqref{eqn:grad}.  Classical PG algorithms use the iterative rule
$\mathbf x_{n+1} = (\mathcal I + \alpha \mathcal F)^{-1}(\mathcal I - \alpha \mathcal G)(\mathbf x_n)$ that converges to the fixed point of \eqref{eqn:grad}. In the linear measurement setting \eqref{eq:linear}, this translates to 
\begin{equation}
\label{eq:proxgrad}
\mathbf x_{n+1} = \underbrace{(\mathcal I + \alpha \mathcal F)^{-1}}_{{\rm prox}_{\alpha} \phi}(\mathbf x_n - \alpha \underbrace{\lambda \mathbf A^H(\mathbf A \mathbf x_n - \mathbf b)}_{\mathcal G(\mathbf x_n)})
\end{equation}
Here, ${\rm prox}_{\alpha}\phi$ is the proximal operator of $\phi$. 
\subsection{Plug and play methods}
The steepest descent update $\mathbf z_n = \mathbf x_n - \alpha \lambda \mathbf A^H(\mathbf A \mathbf x_n - \mathbf b)$ improves the data consistency, while the proximal mapping $\mathbf x_{n+1}= ({\rm prox}_{\alpha} \phi)(\mathbf z_n)$ in \eqref{eq:proxgrad} can be viewed as \emph{denoising} the current solution $\mathbf x_n - \alpha \lambda \mathbf A^H(\mathbf A \mathbf x_n - \mathbf b)$, thus moving the iterate towards the maximum of prior $p(\mathbf x)$. Plug and play methods replace the proximal mapping with off-the-shelf or CNN denoisers $\mathcal H_{\theta}$ \cite{venkatakrishnan2013plug,kamilov2023plug, zhang2018ista, xiang2021fista}:
\begin{equation}\label{pnp-ista}
\mathbf x_{n+1} = \mathcal H_{\theta} \left(\mathbf x_n - \alpha \lambda \mathbf A^H(\mathbf A \mathbf x_n - \mathbf b)\right) = \mathcal T(\mathbf x_n,\theta)
\end{equation}
Here, $\theta$ denotes the learnable parameters. 

There are PnP algorithms that use different optimization algorithms (e.g., ADMM, PG) with convergence guarantees to the fixed point $\mathbf x^* = \mathcal H_{\theta} (\mathbf x^* - \alpha \mathcal G(\mathbf x^*))$ \cite{kamilov2023plug}. The solutions obtained by these approaches often do not have a one-to-one correspondence to the MAP setting in \eqref{eqn:cs}; they may be better understood from the consensus equilibrium setting \cite{buzzard2018plug}. See \cite{kamilov2023plug} for a detailed review of the PnP framework and associated convergence guarantees.

\subsection{Unrolled algorithms}
 Unrolled optimization schemes \cite{gregor2010learning,aggarwal2018modl,hammernik2018learning,schlemper2017deep} aim to learn a CNN denoiser, either as a prior or to replace the proximal mapping as in \eqref{pnp-ista}. A key difference with PnP is in the training strategy. The alternation between the physics-based data consistency (DC) update and the CNN update is unrolled for a finite number of iterations and trained end-to-end, while PnP methods alternate between the DC update and the pre-trained CNN. Unrolled schemes learn the CNN parameters that offer improved reconstruction for a specific sampling operator $\mathbf A$. These schemes obtain a deep network with shared CNN blocks. The parameters $\theta$ of the CNN are optimized to minimize the loss $L = \sum_i \|\tilde{\mathbf x}_{\theta,i}- \mathbf x_i\|_2^2$, where $\mathbf x_i$ are the ground truth images and $\tilde{\mathbf x}_{\theta,i}$ are the output of the deep network. The main challenge of this scheme is the high memory demand of the unrolled algorithm, especially in higher-dimensional (e.g., 3D, 2D + time) settings. This is mainly due to memory required for backpropagation updates scaling linearly with the number of unrolls. While memory-efficient techniques \cite{chen2016training,kellman2020memory} have been proposed, these methods come at the cost of increased computational complexity during training. The choice of minimum number of unrolls to offer good performance is usually ten \cite{aggarwal2018modl, hammernik2018learning}, which is feasible for 2D problems. However, this approach is often infeasible for higher-dimensional applications (3D, 4D, 5D).

\subsection{Deep equilibrium models}
To overcome the challenge associated with unrolled schemes, \cite{gilton2021deep} adapted the elegant DEQ approach introduced in \cite{bai2019deep}. Deep equilibrium models assume that the forward iterations in \eqref{pnp-ista} are run until convergence to the fixed point $\mathbf x^*$ that satisfies $\mathbf x^*=\mathcal T(\mathbf x^*,\theta)$. 
This approach allows one to compute the back-propagation steps using fixed-point iterations \cite{bai2019deep,gilton2021deep} with just one physical layer. 

Unlike unrolled methods, the convergence of the forward and backward iterations are key to maintaining the accuracy of backpropagation in DEQ methods. The convergence of such algorithms are analyzed in \cite{gilton2021deep,ryu2019plug}. In the general case with a full-rank $\mathbf A$ considered in \cite{ryu2019plug}, convergence of the forward-backward splitting (FBS) algorithm in \eqref{pnp-ista} is guaranteed when the Lipschitz constant of $\mathcal H_{\theta}$ satisfies $L[\mathcal H_{\theta}] \leq 2 \mu_{\rm min}/(\mu_{\rm max}-\mu_{\rm min})$, where $\mu_{\rm max}$ and $\mu_{\rm min}$ are the maximum and minimum eigenvalues of $\mathbf A^H \mathbf A$ \cite{ryu2019plug}. In many inverse problems (e.g., parallel MRI with high acceleration), $\mathbf A$ is ill conditioned $\mu_{\rm max}>> \mu_{\rm min}$. In these cases $L[\mathcal H_{\theta}]$ should be close to zero to ensure convergence. Similar issues exist with PnP-ADMM and PnP-PROX, as discussed in \cite{gilton2021deep}.

\section{Monotone Operator Learning}
The main goal of this work is to introduce DEQ algorithms that share the desirable properties of CS algorithms, including uniqueness of the solution, robustness to input perturbations, and guaranteed fast convergence. We constrain $\mathcal F(\mathbf x)$ to be an $m$-monotone ($m>0$) CNN network to achieve these goals. 

\subsection{Monotone Operators}
We constrain the CNN module $\mathcal F$ to be $m$-monotone:

\textbf{Assumption:} The operator $\mathcal F:\mathbb C^M \rightarrow \mathbb C^M$ is $m$-monotone if:
\begin{equation}
	\label{eqn:monotonicity_definition}
	\Re \left(\Big\langle \mathbf x - \mathbf y, \mathcal F(\mathbf x) - \mathcal F(\mathbf y) \Big\rangle\right) \geq m\|\mathbf x- \mathbf y\|_2^2, \hspace{6pt} m>0,
\end{equation}
for all $\mathbf x, \mathbf y \in \mathbb C^M$. Here, $\Re(\cdot)$ denotes the real part. Monotone operators enjoy several desirable properties and have been carefully studied in the context of convex optimization algorithms \cite{ryu2016primer}, mainly due to its following relation with convex priors.
\begin{lem}\cite{sun2019online,sun2021scalable}
	\label{lem0}
	Let $\phi: \mathbb C^M \rightarrow \mathbb R_{+}$ be a proper, continuously differentiable, strongly convex function with $m >0$. Then $\mathcal F = \nabla \phi$ is an $m$-monotone operator. 
\end{lem}
While derivatives of convex priors are monotone, the converse is not true in general. Our results show that the parameter $m$ plays an important role in the convergence, uniqueness, and robustness of the algorithm to perturbations. In many CS applications, $\mathbf A$ often has a large null space, and hence the data-consistency term is not strictly convex. The following result shows that constraining $\mathcal F$ to be $m$-monotone is necessary and sufficient to ensure the uniqueness of the fixed point of \eqref{eqn:grad}.

\begin{prop}
	\label{propv4}
	The fixed point of \eqref{eqn:grad} is unique for a specific $\mathbf b$, iff $\mathcal F$ is $m$-monotone with $m>0$.
\end{prop}

The proof is provided in the Appendix. The following result shows that monotone operators can be represented efficiently as a residual neural network.

\begin{prop}
	\label{resnet}
If $\mathcal H_{\theta}:\mathbb C^M \rightarrow \mathbb C^M$ is a $(1-m)$ Lipschitz function:
\begin{equation}\label{key}
	L[\mathcal H_{\theta}] = (1-m); ~~0<m<1.
\end{equation}
Then, the residual function $\mathcal F:\mathbb C^M \rightarrow \mathbb C^M$ 
	\begin{equation}\label{residual}
		\mathcal F=\mathcal I-\mathcal H_{\theta},
	\end{equation}
is $m$-monotone:
\begin{equation}
\Re\left(\langle \mathcal F(\mathbf x) - \mathcal F(\mathbf y), \mathbf x - \mathbf y \rangle\right) \geq m \|\mathbf x - \mathbf y\|^2_2.
\end{equation}
In addition, the Lipschitz constant of $\mathcal F$ is $2-m$:
	\begin{equation}
		\label{lip-relation}
		\|\mathcal F(\mathbf x) - \mathcal F(\mathbf y)\|_2 \leq (2-m)~ \|\mathbf x-\mathbf y\|_2.
	\end{equation}

\end{prop}

This result allows us to construct a monotone operator as a residual CNN. Because $\mathcal F$ is a score network that predicts the noise, $\mathcal H_{\theta}$ can be viewed as a denoiser. 


\subsection{Proposed iterative algorithm}
We now introduce a novel forward-backward algorithm using $m$-monotone CNN $\mathcal F$. To obtain an algorithm with guaranteed convergence even when $\mathbf A$ is low-rank, we swap $\mathcal F$ and $\mathcal G$ in \eqref{eq:proxgrad}:
\begin{eqnarray}
\label{fixed_pt}
\mathbf x_{n+1} &=& \underbrace{(\mathcal I + \alpha \mathcal G)^{-1}}_{{\rm prox}_{\alpha}\mathcal D}(\mathcal I - \alpha \mathcal F)(\mathbf x_n).
\end{eqnarray}
Therefore, this approach involves a gradient descent to improve the prior, followed by a proximal map of the data term. A similar swapping approach was introduced in \cite{hammernik2020machine} to explain MoDL \cite{aggarwal2018modl}. When $\mathcal F$ is an $m$-monotone operator, the Lipschitz constant of the gradient descent step $(\mathcal I-\alpha \mathcal F)(\mathbf x)$ can be made lower than one as shown in Lemma \ref{lem1}, while that of ${\rm prox}_{\alpha}\mathcal D$ is upper-bounded by one.  
This ensures that the resulting algorithm converges. 

\begin{figure}[h!]%
	\centering
	\subfloat[\centering Iterative Rule]{{\includegraphics[width=0.5\textwidth,keepaspectratio=true,trim={5.6cm 5cm 3.2cm 3.7cm},clip]{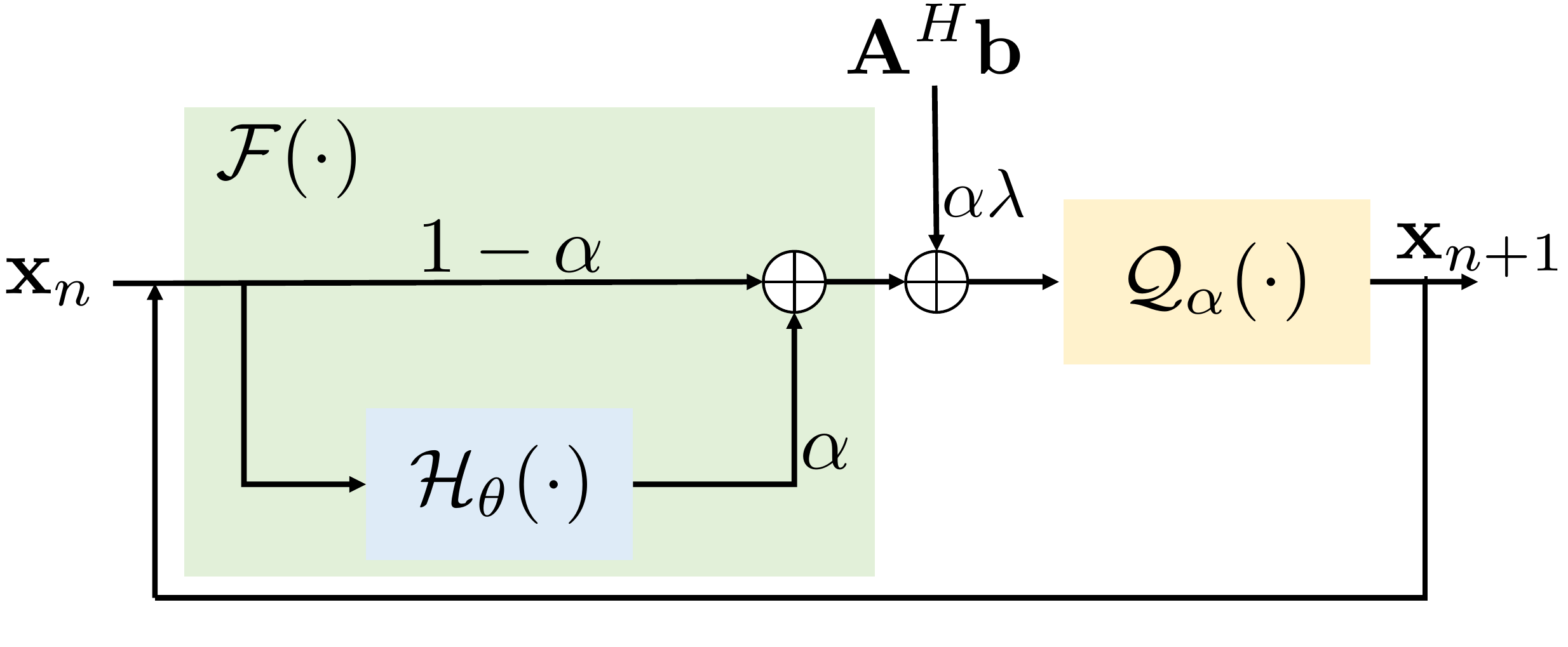}}}%
	\qquad
	\subfloat[\centering CNN architecture]{{\includegraphics[width=0.5\textwidth,keepaspectratio=true,trim={5.6cm 5cm 3.2cm 3.8cm},clip]{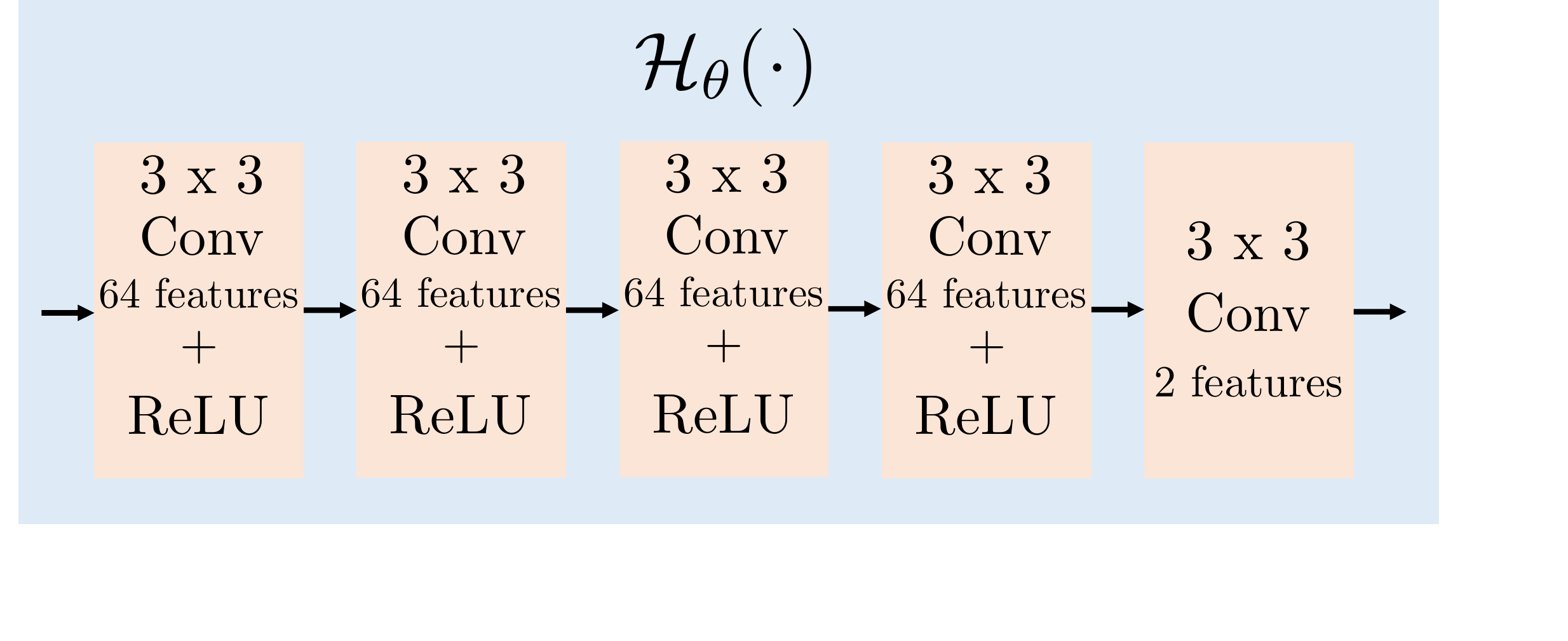}}}%
	\caption{Figure (a) shows fixed-point iterative rule of the proposed MOL algorithm from \eqref{fp} and (b) shows the architecture of the five-layer CNN $\mathcal H_{\theta}(\cdot)$ used in the experiments. When $\alpha = 1$, the approach reduces to the MoDL \cite{aggarwal2018modl}, which was originally introduced for unrolled optimization, or RED \cite{romano2017little}, which was introduced for PnP models.  Our analysis shows that using $\alpha=1$ in the DEQ setting requires highly constrained networks for convergence, which translates to poor performance.}%
	\label{fig:arch}%
\end{figure}

The fixed points of the above relation are equal to the fixed points of \eqref{eqn:cs} for all $\alpha >0$. In the linear setting considered in \eqref{eqn:cs}, suppose we have $\mathbf p = ({\rm prox}_{\alpha} \mathcal D)(\mathbf u)=(\mathcal I+\alpha \mathcal G)^{-1} \mathbf u$, which is the solution of $
\mathbf p+\alpha \underbrace{(\lambda \mathbf A^H(\mathbf A(\mathbf p)- \mathbf b))}_{\mathcal G(\mathbf p)} =  \mathbf u$, or 
\begin{equation}\label{inverse}
({\rm prox}_{\alpha} \mathcal D)(\mathbf u) = \underbrace{\left(\mathbf I+\alpha \lambda \mathbf A^H\mathbf A\right)^{-1}}_{\mathcal Q_{\alpha}} \left(\mathbf u + \alpha \lambda \mathbf A^H \mathbf b\right)
\end{equation}
Combining with \eqref{residual} and \eqref{fixed_pt}, we obtain the proposed MOL algorithm:

\begin{eqnarray}\nonumber
\label{fp}
\mathbf x_{n+1} & = &  \underbrace{(\mathbf I + \alpha \lambda \mathbf A^H \mathbf A)^{-1}\Big((1-\alpha)\mathbf x_n + \alpha \mathcal H_{\theta}(\mathbf x_n) \Big)}_{\mathcal T_{\rm MOL}(\mathbf x_n)} +\\
&&\qquad \underbrace{(\mathbf I + \alpha \lambda \mathbf A^H \mathbf A)^{-1}\Big( \alpha \lambda \mathbf A^H \mathbf b\Big)}_{\mathbf z}  \\
&=& \mathcal T_{\rm MOL}(\mathbf x_n) + \mathbf z
\end{eqnarray}

We show below that this iterative rule will converge to a unique fixed point $\mathbf x^*(\mathbf b)$ specified by $\mathcal T_{\rm MOL}(\mathbf x^*(\mathbf b)) + \mathbf z=\mathbf x^*(\mathbf b)$, which is identical to \eqref{eqn:grad}.

\subsection{Relation to existing algorithms}
\label{alphasection}
We now consider a special case, which has been introduced by other researchers.	When $\alpha=1$, the fixed-point algorithm in the iterative rule in \eqref{fp} can be rewritten as 
 \begin{equation}
 \label{MoDL}
\mathbf x_{n+1}= \left(\mathbf I+ \lambda \mathbf A^H\mathbf A\right)^{-1} \left(\mathcal H_{\theta}(\mathbf x_n)+\lambda \mathbf A^H \mathbf b\right).
\end{equation}
We note that the above update rule is used by multiple algorithms \cite{aggarwal2018modl,hammernik2020machine}, \cite[equation (37)]{romano2017little}. This update rule has been used in fixed-point RED algorithm (see \cite[equation (37)]{romano2017little}). Model-based deep learning (MoDL) \cite{aggarwal2018modl} has trained $\mathcal H_{\theta}$ by unrolling the iterative algorithm with a fixed number of iterations; it did not require the iterative rule to converge. Our analysis shows that the corresponding DEQ algorithms will converge only if the Lipschitz constant of $\mathcal H_{\theta}$ is very low, which would translate to poor performance. The update rule in \eqref{fp} can be viewed as a damped version of MoDL or the fixed-point RED algorithm. As will be seen in our analysis later, the use of the damping factor $\alpha<1$ enables us to relax the constraints on the CNN network $\mathcal H_{\theta}$ that are needed for convergence, which will translate to improved performance. For both MoDL and the proposed fixed-point algorithms, the scalar $\lambda$ is kept trainable.

\section{Theoretical Analysis}

The monotone nature of $\mathcal F$ allows us to characterize the fixed point of the iterative algorithm $\mathbf x_{n+1} = \mathcal T_{\rm MOL} ~(\mathbf x_n) + \mathbf z$ in \eqref{fp}. In particular, we will now analyze the convergence and the robustness of the solution to input perturbations. 

\subsection{Convergence of the algorithm to a fixed point}
The algorithm specified by \eqref{fp} converges if the Lipschitz constant of 
\begin{equation}\label{lipfp}
	\mathcal T_{\rm MOL}(\mathbf x) = \mathcal Q_{\alpha} \left(\mathbf x - \alpha \underbrace{(\mathcal I - \mathcal H_{\theta})}_{\mathcal F} \mathbf x\right) = \mathcal Q_{\alpha} \circ \underbrace{\left(\mathcal I - \alpha\mathcal F \right)}_{\mathcal R}\mathbf x
\end{equation}
is less than one. We will now focus on the composition $\mathcal Q_{\alpha}\circ \mathcal R$. When $\mathbf A^H\mathbf A$ is full-rank, $\mathcal Q_{\alpha}=\left(\mathbf I+\alpha \lambda \mathbf A^H\mathbf A\right)^{-1}$ is a contraction. In many inverse problems including super-resolution and compressed sensing, the Lipschitz constant of $\mathcal Q_{\alpha}$ is $1$. Assuming that $\mathcal F$ is $m$-monotone, we have the following result for $L[\mathcal R]$.
\begin{lem}
\label{lem1}
Let $\mathcal F: \mathbb C^{M}\rightarrow \mathbb C^{M}$ be an $m$-monotone operator. Then, the operator $\mathcal R = (\mathcal I - \alpha \mathcal F)$ has a 
Lipschitz constant of

\begin{equation}
\label{lr}
    L[\mathcal R] \leq \sqrt{1- 2\alpha m+\alpha^2 (2-m)^2 .}
\end{equation}

\end{lem}
From the above relation, we note that $\mathcal R$ is a contraction (i.e., $L[\mathcal R]<1$) when the damping factor $\alpha$ satisfies
\begin{equation}
\label{condition}
   \alpha  < \frac{2 m}{(2-m)^2} = \alpha_{\rm max}.
\end{equation}

\begin{prop}
\label{thmiv3}
Consider the algorithm specified by \eqref{fp}, where $\mathcal F$ is an $m$-monotone operator. Assume that \eqref{fp} has a fixed point specified by $\mathbf x^*(\mathbf b)$. Then, 
    \begin{equation}
    \|\mathbf x_{n}-\mathbf x^*(\mathbf b)\|_2 \leq (L[\mathcal T_{\rm MOL}])^{n} ~\|\mathbf x_0-\mathbf x^*(\mathbf b)\|_2,
\end{equation}
where $L[\mathcal T_{\rm MOL}] = \frac{1}{(1+\lambda \mu_{\rm min})}~L[\mathcal R]$. Here, $\mu_{\rm min}$ is the minimum eigenvalue of $\mathbf A^H\mathbf A$ and $L[\mathcal R]$ is specified by \eqref{lr}.
\end{prop}

We note that $\mathcal T_{\rm MOL}$ being a contraction translates to geometric convergence with a factor of $L[\mathcal T_{\rm MOL}]$; this is faster than the sublinear convergence rates \cite{sun2019online,sun2021scalable} available for ISTA \cite{daubechies2004iterative} and ADMM \cite{yang2010fast} in the CS setting ( $\mu_{\rm min} = 0)$.

\subsection{Benefit of damping parameter $\alpha$ in the MOL algorithm \eqref{fp}}

We note from Section \ref{alphasection} that the algorithms \cite{aggarwal2018modl,hammernik2020machine,romano2017little}
correspond to the special case of $\alpha = 1$. Setting $\alpha=1$ in \eqref{condition}, we see from Lemma IV.1 and \eqref{condition} that the DEQ algorithm will converge if 
\begin{equation}
    m \geq 3-\sqrt{5}=0.76
\end{equation}
or $L[\mathcal H_{\theta}]<0.24$. As discussed previously, the denoising ability of a network is dependent on its Lipschitz bound; a smaller $L[\mathcal H_{\theta}]$ bound translates to poor performance of the resulting MOL algorithm. 
The use of the damping factor $\alpha <1$ allows us to use denoising networks $\mathcal H_{\theta}$ with larger Lipschitz bounds and hence improved denoising performance. For instance, if we choose $m=0.1; L[\mathcal H_{\theta}]=0.9$, from \eqref{condition}, the algorithm will converge if $\alpha < 0.055$.

\subsection{Robustness of the solution to input perturbation}
\label{rob_perf_trade_off}
The following result shows that the robustness of the proposed algorithm is dependent on $m$ or, equivalently, $L[\mathcal H_{\theta}]$. We note that the link between the Lipschitz bound on the network and sensitivity to perturbations is straightforward in a direct inversion scheme (e.g., UNET). By contrast, such relations are not available in the context of DEQ-based image recovery algorithms, to the best of our knowledge. 
\begin{prop}
\label{robustness}
Consider $\mathbf z_1$ and $\mathbf z_2$ to be measurements with $\boldsymbol\delta =\mathbf z_2-\mathbf z_1$ as the perturbation. Let the corresponding outputs of the MOL algorithm be $\mathbf x^*(\mathbf z_1)$ and $\mathbf x^*(\mathbf z_2)$, respectively, with $\Delta= \mathbf x^*(\mathbf z_2)-\mathbf x^*(\mathbf z_1)$ as the output perturbation,
\begin{eqnarray}\nonumber
\|\Delta\|_2
&\leq&\frac{\alpha \lambda/(1+\lambda \mu_{\rm min}) }{1-\sqrt{1 - 2\alpha m + \alpha^2(2-m)^2}}~\|\boldsymbol \delta\|_2.\\\label{rob_eqn}
\end{eqnarray}
\end{prop}

\begin{figure}[t!]
	\centering
	\includegraphics[width=0.4\textwidth,keepaspectratio=true]{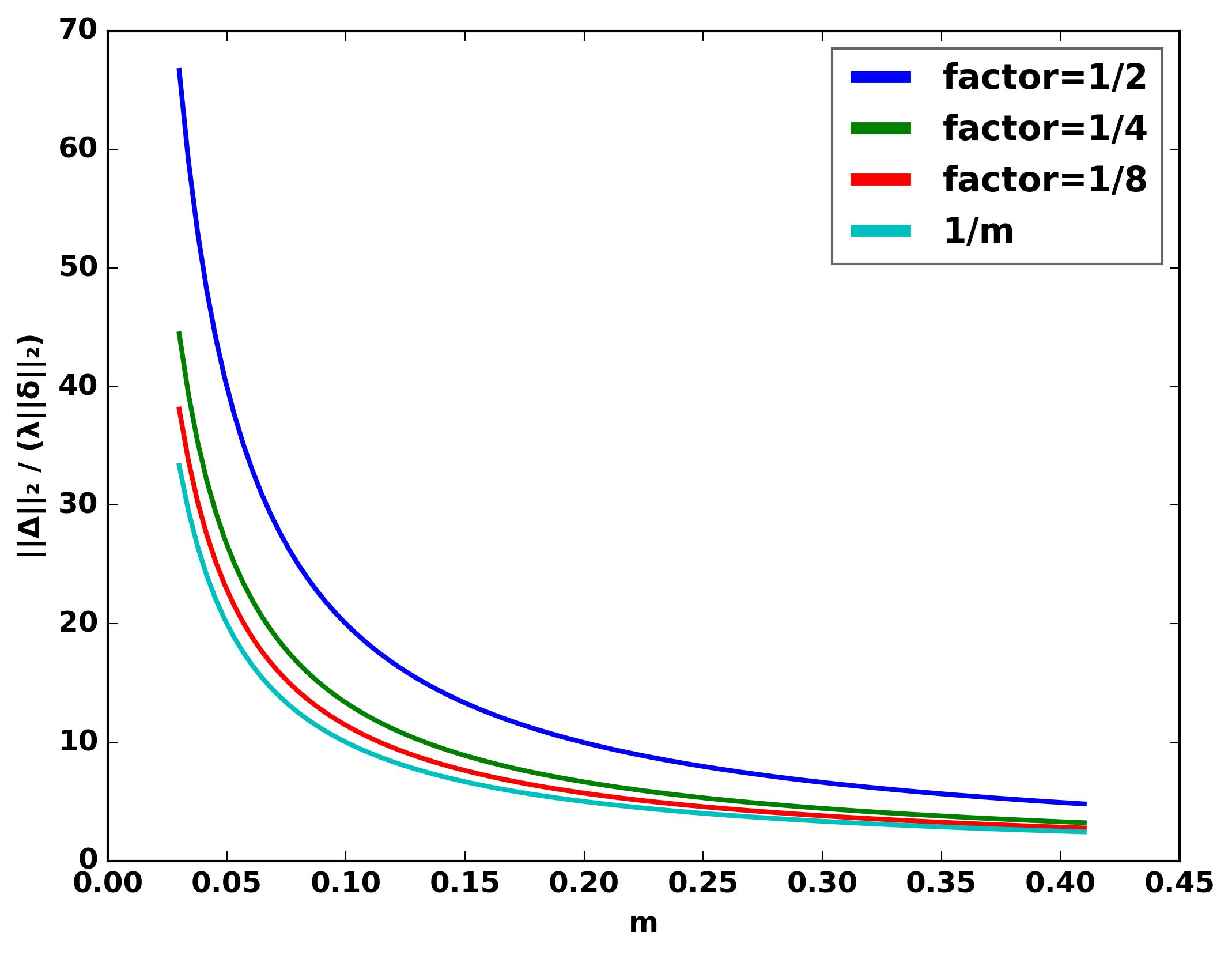}
	\caption{Plot of $\frac{\alpha }{1-\sqrt{1 - 2\alpha m + \alpha^2(2-m)^2}}$ in \eqref{robustness} vs $m$. Here, we choose $\alpha = {\rm factor}\times \alpha_{\rm max}$, where $\alpha_{\rm max}$ is specified by (19) in the paper. We note that all the curves roughly decay with $m$ with a $1/m$ decay rate. As $\alpha \rightarrow 0$ or equivalently low values of factor, the curves approach $1/m$. }
	\label{plot}
\end{figure}

The above result shows that the norm of the perturbation in the reconstructed images is linearly dependent on the norm of the input perturbations. The constant factor is a function of the monotonicity parameter $m$ and the step size $\alpha$ in Fig. \ref{plot}, where we plot the constant term without the $\lambda$ parameter for different values of $m$. The plots show that the constant decreases with $m$ roughly at $1/m$ rate. We note that the algorithm will converge to the same fixed point as long as the damping parameter $\alpha$ satisfies the condition \eqref{condition}. Hence, we consider the case with small damping parameter $\alpha \rightarrow 0$ and set $\mu_{\rm min}=0$ corresponding to the CS and super-resolution settings, when we obtain a simpler expression:
\begin{equation}\label{simpler}
	\lim_{\alpha\rightarrow 0}\|\Delta\|_2 \leq \frac{\lambda}{m} \|\boldsymbol\delta\|_2.
\end{equation}
The above results show that the robustness of the algorithm is fundamentally related to $m$; a higher value of $m$ translates to a more robust algorithm. However, note that the Lipschitz constant of the denoiser $\mathcal H_{\theta}$ is specified by $L[\mathcal H_{\theta}]=1-m; m>0$. We need to choose a denoising network with a lower Lipschitz constant, which translates to lower performance, to make the resulting algorithm more robust to perturbations. There is a trade-off between robustness and performance of the algorithm, controlled by either the parameter $m$ or the Lipschitz constant $L[\mathcal H_{\theta}]$.

\section{Implementation Details}

\subsection{DEQ: forward and backward propagation}
\label{molfwdbwd}
During training and inference, we use the forward iteration rule, $\mathbf x_{n+1} = \mathcal T_{\rm MOL}(\mathbf x_n)+ \mathbf z$.
We terminate the algorithm at the $n^{th}$ iteration $\mathbf x_n$ if it satisfies,
	\begin{equation}
		\label{termination}
		e_n = \frac{\|\mathbf x_{n} - \mathbf x_{n-1}\|_2}{\|\mathbf x_{n-1}\|_2} \leq \kappa .    
	\end{equation} 
We set $\kappa=1\times 10^{-4}$ for the experiments. Please see the pseudo-code for forward propagation in Algorithm 1 of the supplementary material. We denote the fixed point of the algorithm as $\mathbf x^{*}(\mathbf b)$, such that $\mathbf x^{*}(\mathbf b) \approx \mathcal T_{\rm MOL}\left(\mathbf x^{*}(\mathbf b)\right)+\mathbf z$. DEQ schemes \cite{gilton2021deep} rely on fixed-point iterations for back-propagating the gradients. The details are shown in Algorithm 1 and 2 in the supplementary material. The iterations are evaluated until convergence, using similar termination conditions as in  \eqref{termination}.

\subsection{Implementation of the monotone CNN operator}
We note from \eqref{resnet} that a monotone $\mathcal F = \mathcal I - \mathcal H_{\theta}$ can be learned by constraining the Lipschitz constant of the denoiser network $\mathcal H_{\theta}$. We propose two different implementations of the MOL algorithm, which differs in the way the Lipschitz constraint is implemented. 

\subsubsection{Spectral normalization}
Similar to \cite{miyato2018spectral}, we use normalization of the spectral norm of the CNN layers to constrain $L[\mathcal H_{\theta}]$. In particular, we bound the spectral norm of each layer to $\sqrt[s]{L[\mathcal H_{\theta}]}$, where $s$ is the number of layers. We term this version of MOL as MOL-SN (MOL-spectral normalization). This approach can guarantee $\mathcal H_{\theta}$ to be a contraction, and hence the guarantees are satisfied exactly. However, the product of the spectral norms of the individual layers is a conservative estimate of the Lipschitz constant of the network. As shown by our experiments, the use of spectral normalization in our setting (MOL-SN) translates to lower performance. Another challenge with the spectral normalization approach is that it restricts the type of networks that can be used; architectures with skipped connections cannot be used with this strategy. We note that spectral normalization is indeed a conservative bound for the Lipschitz constant and hence may over constrain the network, translating to lower performance.  

\subsubsection{Approximating monotone constraint using a Lipschitz penalty}
Motivated by \cite{bungert2021clip}, we propose to train the MOL algorithm using a training loss which minimizes a constrained optimization problem. In \cite{pesquet2021learning}, authors use Jacobian regularization instead to learn a contractive network. The estimation of the Lipschitz constant of $\mathcal H_{\theta}$ is posed as a maximization problem \cite{bungert2021clip}:
\begin{equation}
	\label{Lipschitz}
	l\left[\mathcal H_{\theta}\right] = \max_{\mathbf x \in S}\overbrace{\sup_{\boldsymbol\eta} \underbrace{\frac{\|\mathcal H_{\theta}(\mathbf x + \boldsymbol\eta) -\mathcal H_{\theta}(\mathbf x)\|_2^2}{\|\boldsymbol\eta\|_2^2}}_{p(\mathbf x, \boldsymbol\eta)}}^{P(\mathbf x)}
\end{equation} 
We denote the estimated Lipschitz constant as $l\left[\mathcal H_{\theta}\right]$ to differentiate it from the true Lipschitz constant $L\left[\mathcal H_{\theta}\right]$. Here, $\boldsymbol\eta$ is a perturbation, and $S$ is the set of training data samples. We note that this estimate is less conservative than the one using spectral normalization. However, this is only an approximation of the true Lipschitz constant, even though our experiments show that the use of this estimate can indeed result in algorithms with convergence and robustness as predicted by the theory. We note that several researchers have recently introduced tighter estimates for the Lipschitz constant \cite{latorre2020lipschitz,jordan2020exactly}, and they could be used to replace the above estimate. The theoretical results derived in the earlier sections will still hold, irrespective of the specific choice of the Lipschitz estimation strategy. We initialize $\eta$ by a small random vector, which is then updated using steepest ascent. 
It is solved using a log-barrier approach which constrains the estimated Lipschitz of the CNN below a threshold value. The total training loss is a linear combination of the log-barrier term and the supervised mean squared error (MSE) loss. We call this method as MOL-LR (MOL-Lipschitz Regularized). 

\subsection{Training the MOL-LR algorithm}

In the supervised learning setting, we propose to minimize 
\begin{eqnarray}\nonumber
	\label{supervisedloss}
	\mathcal C(\theta) &=& \underbrace{\sum_{i=0}^{N_t} \|\mathbf x_i^* - \mathbf x_{i}\|_2^2}_{\mathcal C}~~ \mbox{such that }\\
	&&~~~~ \underbrace{P\left(\mathbf x_i^*\right)}_{\text{Local Lipschitz estimate}}\leq T; i=0,..,N_t
\end{eqnarray}
Here, the threshold is selected as $T = 1-m$ and $P(\mathbf x)$ is indicated in \eqref{Lipschitz}. The above loss function is minimized with respect to parameters $\theta$ of the CNN $\mathcal H_{\theta}$.  $\mathbf x_i^*$ is a fixed point of \eqref{eqn:grad} described in Section \ref{molfwdbwd}, which is dependent on the CNN parameters $\theta$. $\mathbf x_i; i=0,..,N_t$ and $\mathbf b_i, i=0,..,N_t$ are the ground truth images in the training dataset and the corresponding under-sampled measurements, respectively. We solve the above constrained optimization scheme by using a log-barrier approach:

\begin{equation}
	\label{finalloss}
	\theta^* = \arg \min_{\theta}~\underbrace{\sum_{i=0}^{N_t} \Big(\|\mathbf x_i^* - \mathbf x_{i}\|_2^2 -  \beta \log \left(T - P\left(\mathbf x_i^*\right)\right)\Big)}_{C_i} .
\end{equation}

Here $\beta$ is a parameter that decays with training epochs similar to conventional log-barrier methods. This optimization strategy ensures that the estimate does not exceed $T$. For implementation purposes, we evaluate the worst-case perturbations $\boldsymbol\eta_i$ for each $\mathbf x_i^*$ by maximizing \eqref{Lipschitz} at each epoch. These perturbations are then assumed to be fixed to evaluate the above loss function, which is used to optimize $\theta$.  
The training algorithm is illustrated in the pseudo-code shown in Algorithm \ref{training}, which is illustrated for a batch size of a single image and gradient descent for simplicity.
\begin{algorithm}
	\caption{: Training: input=training data $\mathbf x_i; i=1,..,N_t$}
	 \label{training}
	\begin{algorithmic}[1]
		\For {ep = $1,2,\ldots$ }
			\For {$i=1,2,\ldots, N_t$ }
				\State Determine $\mathbf x_i^*$ using DEQ forward iterations
				\State Determine $\boldsymbol\eta_i^* = \arg \max_{\boldsymbol\eta} p(\mathbf x_i^*,\boldsymbol\eta)$
				\State $C_i = \|\mathbf x_i-\mathbf x_i^*\|^2 - \frac{1}{m} \log(T-p(\mathcal H_{\theta}(\mathbf x_i^*),\boldsymbol\eta_i^*))\ $
            \State Determine $\nabla_{\theta} C_i$ using DEQ backward iterations
    \State $\theta \leftarrow \theta - \gamma \nabla_{\theta} C_i$, where $C_i$ is the loss in \eqref{finalloss}
			\EndFor
		\EndFor
	\end{algorithmic} 
\end{algorithm}

\subsection{Unrolled algorithms used for comparison}
We compare the proposed MOL algorithm against SENSE \cite{pruessmann1999sense}, MoDL \cite{aggarwal2018modl}, ADMM-Net \cite{sun2016deep}, DE-GRAD \cite{gilton2021deep}, and UNET \cite{ronneberger2015u}. SENSE is a CS-based approach that uses a forward model consisting of coil sensitivity weighting and undersampled Fourier transform. MoDL and ADMM-Net are unrolled deep learning algorithms, which alternate between the DC step and the CNN-based denoising step. Both approaches are trained in an end-to-end fashion for 10 iterations. DE-GRAD is a deep equilibrium network, where we use spectral normalization as described in \cite{gilton2021deep}. UNET is a direct inversion approach, which uses a CNN without any DC steps. We choose five-layer CNNs for all the unrolled deep-learning and DEQ based algorithms used for comparisons. The CNN architecture is shown in Fig. \ref{fig:arch} (b).

We consider two versions of MOL: MOL-SN, which relies on spectral normalization during training to constrain the Lipschitz constant of the overall network, and MOL-LR, which consists of an additional loss term computing the Lipschitz constant of the CNN. 
We also consider Lipschitz regularized versions of UNET, ADMM-Net, and MoDL for robustness experiments, and those are denoted by UNET-LR, ADMM-Net-LR, and MoDL-LR, respectively. The Lipschitz of the CNNs in these methods is regularized by the proposed training strategy in \eqref{finalloss}. For 2D+time experiments on cardiac data, we compare a 2D+time version of MOL-LR (with 3D convolutions) against the 2D MoDL (with 2D convolutions).   

\subsection{Architecture of the CNNs and training details}
The MOL architecture is shown in Fig. \ref{fig:arch}. In our 2D experiments, the CNN $\mathcal H_{\theta}$ consists of five 2D convolution layers, each followed by rectified linear unit (ReLU) non-linearity, except for the last layer. The convolution layers consist of 64 filters with 3x3 kernels. The parameter $\lambda$ in \eqref{fp}, weighing the DC term, is kept trainable. A SENSE reconstruction with $\lambda_0 = 100$ is performed initially on the undersampled image $\mathbf A^H\mathbf b$ to initialize the MOL network as $\mathbf x_0$. A similar approach is used for the other deep learning networks (MoDL, ADMM-Net, UNET) to ensure fair comparisons. We share the CNN weights across the iterations for all the unrolled deep learning algorithms (MoDL, ADMM-Net). We use a full-size UNET, consisting of four layers of pooling and unpooling. Note that the number of trainable parameters in the chosen UNET is at least twice the number of parameters in five-layer CNNs. In 2D+time experiments, a five-layer CNN is chosen; it is similar to the 2D case, with the exception of 3D convolution layers instead of 2D.  We set $\alpha = 0.055$, which corresponds to $m=0.1$ or  $L[\mathcal H_{\theta}]=0.9$ for all MOL algorithms. We keep this variable fixed and non-trainable because this parameter is chosen based on \eqref{condition}, which depends on the bound $T=1-m$. We show in Fig. \ref{fig:brain_train_plots} that the MOL algorithm with $\alpha=1$, which is the DEQ extension of MoDL \cite{aggarwal2018modl} and RED \cite{romano2017little}, diverges.

All the trainings are performed on a 16 GB NVIDIA P100 GPU. The CNN weights are Xavier initialized and trained using an Adam optimizer for 200 epochs. The learning rates for updating CNN weights and $\lambda$ are chosen empirically as $10^{-4}$ and $1.0$, respectively. MoDL and ADMM-Net are unrolled for 10 iterations; MOL, on the other hand, consumes memory equivalent to a single iteration. All the methods are implemented in PyTorch. During inference, the reconstruction results are quantitatively evaluated in terms of the Structural Similarity Index (SSIM) \cite{wang2004image} and the Peak Signal-to-Noise Ratio (PSNR). 

\subsection{Computing worst-case (adversarial) perturbations}
We determine the robustness of the networks to Gaussian as well as worst-case perturbations. We determine the worst-case perturbation $\boldsymbol\gamma$ by solving the following optimization problem:
\begin{equation}
	\boldsymbol\gamma^*= \max_{\boldsymbol\gamma;~ \|\boldsymbol\gamma\|_2<\epsilon\cdot\|\mathbf b\|_2} \underbrace{\|\mathbf x^*\left(\mathbf b + \boldsymbol\gamma\right) - \mathbf x^*(\mathbf b)\|_2^2}_{U(\boldsymbol\gamma)}
\end{equation} 
Here, $\gamma$ is solved using a projected gradient algorithm; we alternate between gradient ascent steps and renormalization of $\boldsymbol\gamma$ to satisfy the constraint $\|\boldsymbol\gamma\|_2<\epsilon\cdot\|\mathbf b\|_2$.  For MOL, we use the fixed-point iterations described in Section \ref{molfwdbwd} to compute the gradient $\nabla_{\boldsymbol\gamma}U(\boldsymbol\gamma)$. We note that the fixed-point iterations for back-propagation are accurate as long as the forward and backward iterations converge. We track the maximum number of iterations and the termination criterion \eqref{termination} to ensure that the iterations converge. In this work we relied on an $\ell_2$ norm on the perturbations, while $\ell_{\infty}$ constraints have also been used in the literature  \cite{antun2020instabilities}. 

\subsection{2D Brain and knee datasets}

We used the 2D multi-coil brain data from the publicly available Calgary-Campinas Public (CCP) Dataset \cite{souza2018open}. The dataset consists of twelve-coil T1-weighted brain data from 117 healthy subjects, collected on a 3.0 Tesla MRI scanner. The scan parameters are: TR (repetition time)/TE (echo time)/TI (inversion time) = 6.3 ms/2.6ms/650 ms or TR/TE/TI = 7.4ms/3.1ms/400ms. Matrix sizes are 256x208x170/180 with 256, 208, and 170/180 being the readout, phase encoding, and slice encoding directions, respectively. For the experiments, we choose subjects with fully sampled data (67 out of 117) and split them into training (45), validation (2), and testing (20) sets. The k-space measurements are retrospectively undersampled along the phase and slice encoding directions using a four-fold 2D non-uniform variable-density mask.

We also perform experiments on the multi-channel knee MRI datasets from the fastMRI challenge \cite{zbontar2018fastmri}. It consists of 15-coil coronal proton-density weighted knee data with or without fat suppression. The sequence parameters were: matrix size 320 x 320, in-plane resolution 0.5mm x 0.5mm, slice
thickness 3mm, and repetition time (TR)
varying from 2200 to 3000 ms, and echo time (TE) between 27 and 34 ms. We use the k-space measurements from 50 subjects for training, 5 for validation, and 20 for testing, respectively. The data is retrospectively undersampled along the phase-encoding direction, for four-fold, using a 1D non-uniform variable density mask. In another set of experiments, we consider four-fold undersampling using 1D uniform mask.  

\subsection{Cardiac MRI datasets}

Compressed sensing and low-rank methods have been extensively used to reduce the breath-hold duration in cardiac cine \cite{vincenti2014compressed,zhao2010psf}.  Several authors have introduced unrolled algorithms for cardiac cine MRI acceleration. For instance, one of the initial works considered the independent recovery of 2D images using unrolled methods \cite{schlemper2017deep}, together with data sharing. More recent works \cite{sandino2021accelerating,zucker2021free} rely on a 15-iteration unrolled scheme, where they used separable (2+1)-D spatio-temporal convolutions to keep the memory demand manageable.

In this work, we show the preliminary utility of the proposed MOL approach with 3D CNN to accelerate cardiac cine MRI. We used the multi-coil cardiac data from the open-source OCMR Dataset \cite{chen2020ocmr}. We note that the high memory demand often restricts the training of unrolled algorithms such as MoDL \cite{aggarwal2018modl} in the 2D+time setting. We chose data from 1.5 Tesla scans, which consists of a series of short-axis, long-axis, or four-chamber views. We use fifteen subjects for training, two for validation, and eight for testing. Each dataset consists of 20-25 time-frames per slice, with 1-3 slices per subject. We retrospectively undersample the k-t space data using a 1D non-uniform variable-density mask along the phase encode direction. In these experiments, we compare the MOL algorithm using a 3D network against the MoDL algorithm \cite{aggarwal2018modl} using a 2D network. The MOL training is performed on 2D+time datasets. On the other hand, the MoDL algorithm \cite{aggarwal2018modl} processes each of the time-frames independently, and hence is not capable of exploiting inter-frame dependencies.

\section{Experiments and Results}
The proposed method can handle wide range of inverse problems. We showcase its application in parallel MRI recovery from non-uniform and uniform undersampled acquisition settings in the subsequent sections. In addition, we also demonstrate it on image super-resolution problems. The super-resolution experiments and results are discussed at Section III in the supplementary material.

\subsection{Characterization of the models}
\begin{figure*}[t!]
	\centering
	\includegraphics[keepaspectratio=true,trim={1.7cm 11.8cm 6.2cm 12cm},clip,width=\textwidth]{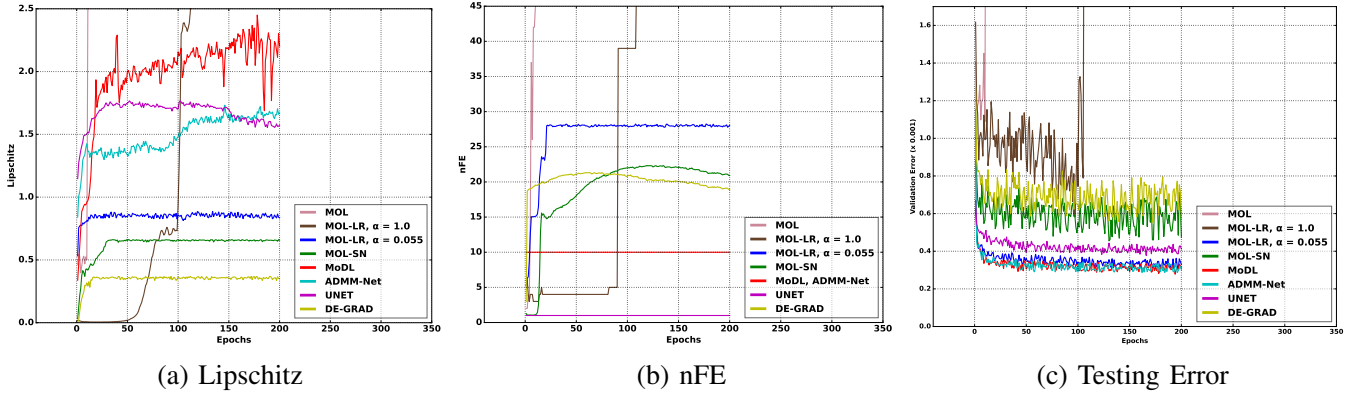}
	\caption{Convergence behaviour of MOL and other algorithms. MOL corresponds to the proposed algorithm without any Lipschitz constraint, MOL-LR, $\alpha=1$ is the DEQ version of MoDL/RED \cite{aggarwal2018modl,romano2017little} with Lipschitz constraint, and MOL-LR, $\alpha=0.055$ is the proposed scheme. 
 Graphs are plotted with respect to epochs during training. (a) shows the evolution of the Lipschitz constant of the CNN module. (b) plots the number of iterations used in the algorithm. (c) plots the testing error on the validation datasets used during training. For MoDL and ADMM-Net, we have ten unrolls, and UNET consists of a single CNN. The DEQ forward and backward algorithms are run until the difference between the subsequent terms satisfies the convergence criterion from \eqref{termination}. We note that the MOL and MOL-LR with $\alpha=1$ diverges as the training proceeds, as predicted by the theory. By contrast, the proposed MOL-LR scheme converges with $\approx 28$ iterations or number of function evaluations (nFEs). We note that MOL-LR requires more forward iterations to converge than MOL-SN, mainly because of the higher Lipschitz constant of $\mathcal H_{\theta}$. }
	\label{fig:brain_train_plots}
\end{figure*}
In Fig. \ref{fig:brain_train_plots}, we show the characteristics of the different models during training. Fig. \ref{fig:brain_train_plots}.(a) shows plots of Lipschitz constant against epochs for different methods. We note that the computed Lipschitz constant $l[\mathcal H_{\theta}]$ of the MOL-SN and DE-GRAD schemes that use spectral normalization stays around 0.7 and 0.4, respectively, which translates to lower performance seen from the testing error curves in (c). By contrast, the unrolled MoDL and ADMM-Net, and UNET, have no Lipschitz constraints and, therefore, have more flexibility in CNN weight updates. It is observed that the estimated Lipschitz constants of these networks often exceed 1. The proposed MOL-LR algorithm maintains the Lipschitz constant less than 0.9. The slightly lower performance of MOL-LR compared to that of other unrolled methods can be attributed to its lower Lipschitz constant.

The plot in Fig. \ref{fig:brain_train_plots}.(b) shows the number of iterations of \eqref{fp} (equivalently, the number of CNN function evaluations, denoted by nFE) needed to converge to the fixed point with a precision of $\kappa=1\times10^{-4}$. As expected, MOL-SN and DE-GRAD, which have lower Lipschitz constants and hence higher $m$, converge more rapidly than MOL-LR, which has a lower value of $m$. We note that the MOL-LR for $\alpha=1$ algorithm is essentially the MoDL algorithm \cite{aggarwal2018modl} used in the DEQ setting. As predicted by theory, we note that this algorithm fails to converge, evidenced by the number of iterations (nFE) increasing rapidly around 100 epochs. Similarly, the MOL algorithm without Lipschitz constraints also diverges around 10 epochs. The proposed MOL-LR with $\alpha=0.055$ converges at approximately nFE = 28.

\subsection{Performance comparison in the parallel MRI setting}

\begin{figure*}[h!]
\centering
\includegraphics[width=\textwidth,keepaspectratio=true,trim={1.8cm 9.7cm 0.4cm 9.9cm},clip]{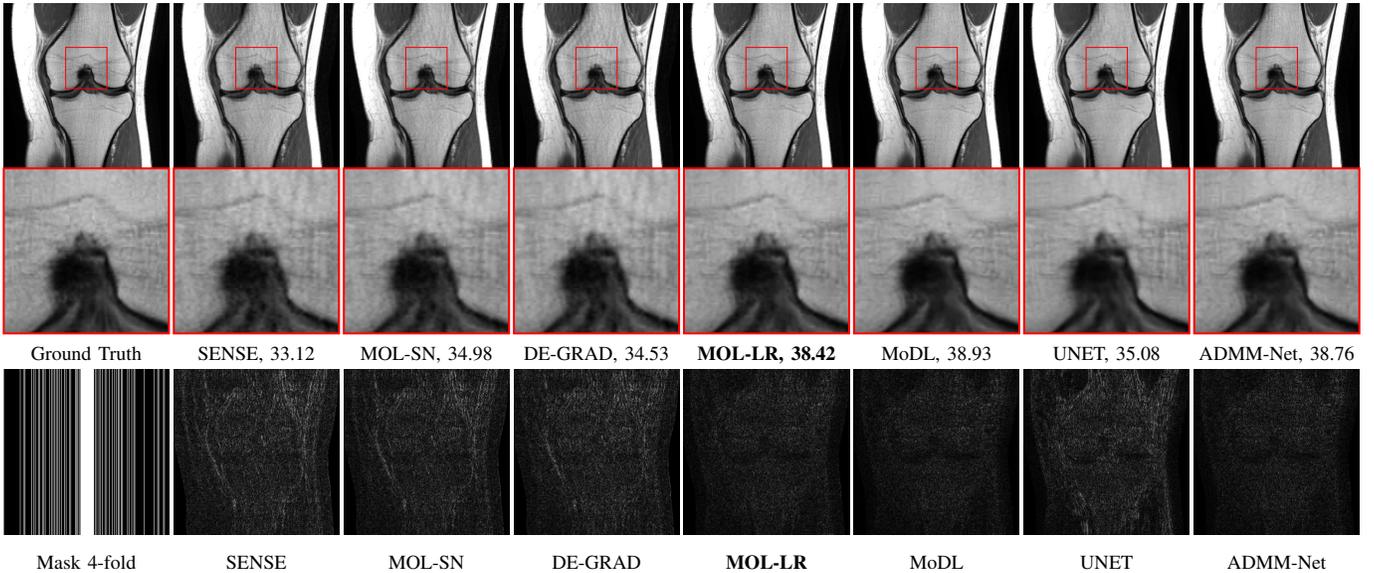}
\caption{Reconstruction results of 4x accelerated multi-channel fastMRI knee data with variable density sampling. PSNR (dB) values are reported for each case. The image in the first row of the first column was undersampled along the phase-encoding direction using a Cartesian 1D non-uniform variable-density mask as shown in the second row of the first column. The top row shows reconstructions (magnitude images), while the bottom row shows corresponding error images. We note that the quality of the \textbf{MOL-LR} reconstructions is comparable to unrolled methods MoDL and ADMM-Net. MOL-SN and DE-GRAD show significantly lower performance due to spectral normalization of weights, resulting in stricter bounds on the Lipschitz constant of its CNN.}
\label{fig:knee_perf_compare}
\end{figure*}

\begin{table}[ht!]
\fontsize{7}{10}
\selectfont
\centering
\begin{tabular}{|c|ccc|}
\hline
& \multicolumn{3}{c|}{Four-fold Knee MRI} \\ \hline
Methods & PSNR & SSIM & Run-time\\ \hline 
SENSE  & 33.04 $\pm$ 1.44 & 0.986 $\pm$ 0.023 & \textbf{0.07s}\\
MOL-SN & 34.86 $\pm$ 1.26 & 0.987 $\pm$ 0.019 & 0.32s\\
DE-GRAD & 34.47 $\pm$ 1.39 & 0.987 $\pm$ 0.021 & 0.26s\\
MOL-LR & 38.34 $\pm$ 0.83 & \textbf{0.993 $\pm$ 0.005} & 0.47s\\ 
MoDL & \textbf{38.74 $\pm$ 0.77} & \textbf{0.993 $\pm$ 0.005} & 0.19s\\
MoDL-LR & 36.53 $\pm$ 1.01 & 0.990 $\pm$ 0.010 & 0.19s\\
ADMM-Net & 38.63 $\pm$ 0.78 & \textbf{0.993 $\pm$ 0.005} & 0.20s\\
ADMM-Net-LR & 36.45 $\pm$ 1.02 & 0.990 $\pm$ 0.010 & 0.20s\\
UNET & 35.12 $\pm$ 1.19  & 0.988 $\pm$ 0.013 & 0.08s\\
UNET-LR & 33.76 $\pm$ 1.40 & 0.986 $\pm$ 0.022 & 0.08s\\ \hline
\end{tabular}

\vspace{1em}
\caption{Quantitative comparisons on 2D datasets with 4-fold undersampling using Cartesian 1D non-uniform variable-density mask. PSNR in dB, SSIM, and mean run-time per slice in seconds are reported. The PSNR and SSIM values are in mean $\pm$ standard deviation format.}
\label{tab:perf_comp_tab} 
\end{table}

\begin{table}[ht!]
\fontsize{7}{10}
\selectfont
\centering

\begin{tabular}{|c|cc|}
\hline
& \multicolumn{2}{c|}{Four-fold Knee MRI} \\ \hline
Methods & PSNR & SSIM \\ \hline 
SENSE  & 32.76 $\pm$ 1.56 & 0.985 $\pm$ 0.026 \\
MOL-LR & 38.03 $\pm$ 0.85 & 0.992 $\pm$ 0.007 \\ 
MoDL & \textbf{38.62 $\pm$ 0.80} & \textbf{0.992 $\pm$ 0.006} \\ \hline
\end{tabular}

\vspace{1em}
\caption{Quantitative comparisons on 2D datasets with 4-fold undersampling using Cartesian 1D uniform mask. PSNR in dB and SSIM values are reported. The format is mean $\pm$ standard deviation.}
\label{tab:perf_comp_uniform_tab} 
\end{table}

\begin{figure}[h!]
\centering
\includegraphics[width=0.98\linewidth,keepaspectratio=true,trim={1.8cm 9.7cm 10.1cm 9.9cm},clip]{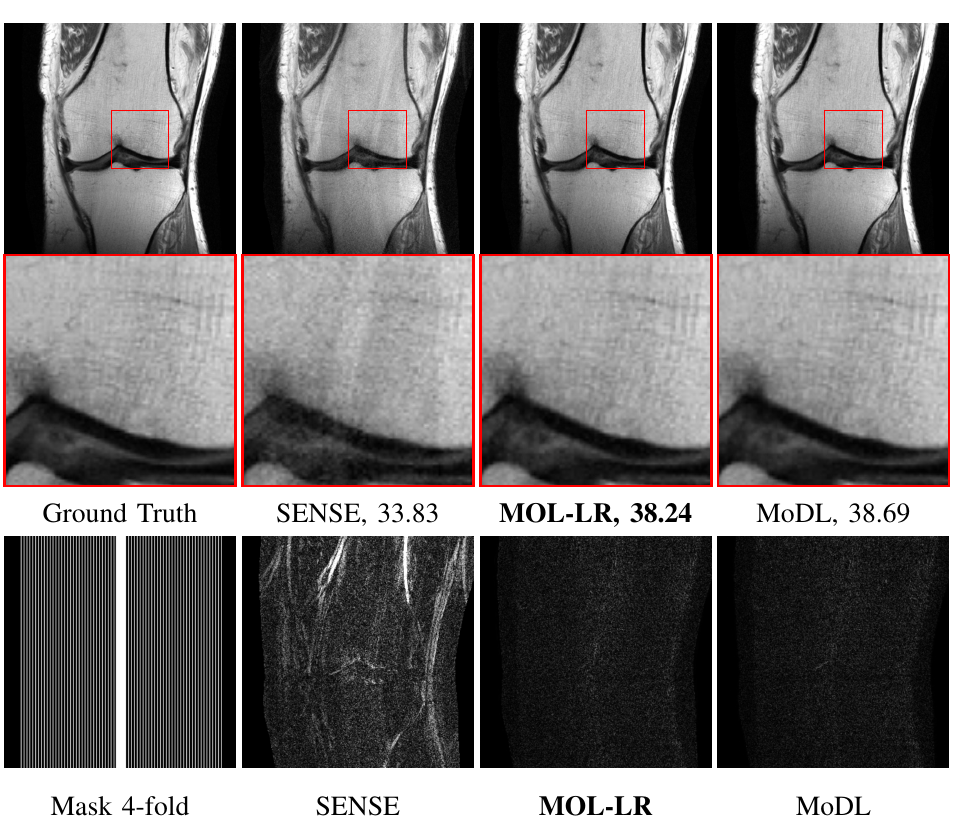}
\caption{Reconstruction results of 4x accelerated knee data with uniform undersampling. PSNR (dB) values are reported for each case. The image in the first row and column was undersampled along phase encoding direction using a Cartesian 1D uniform mask as shown in the first column, second row. The top row shows reconstructions (magnitude images), while the bottom row shows corresponding error images. We note that the quality of the \textbf{MOL-LR} reconstruction is comparable to the unrolled method MoDL. SENSE reconstruction has visible aliasing and performs significantly poorer in this setting.}
\label{fig:knee_uniform}
\end{figure}

\begin{figure*}[!ht]
	\centering
	\includegraphics[scale=0.9,keepaspectratio=true,trim={1.8cm 5.1cm 0.5cm 5cm},clip]{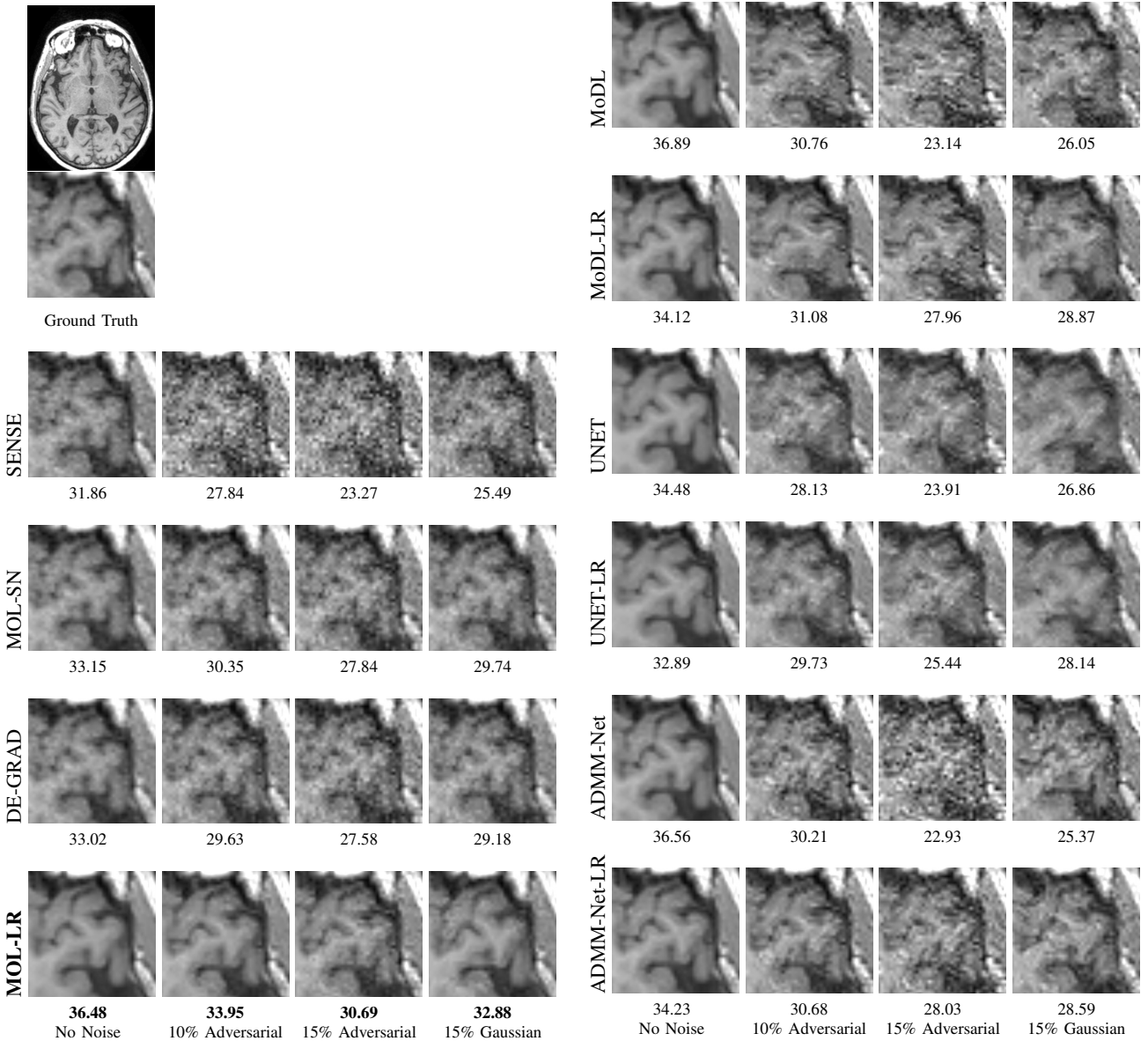}
	\caption{Sensitivity of the algorithms to input perturbations: The rows correspond to reconstructed images from 4x accelerated multi-channel brain data using different methods. The data was undersampled using a Cartesian 2D non-uniform variable-density mask. The columns correspond to recovery from noiseless, worst-case (adversarial) perturbation whose norm is 10\% and 15\% of the measured data, and Gaussian noise whose norm is also 15\% of the measured data, respectively. The PSNR (dB) values of the reconstructed images are reported for each case. We observe that the performance of the MOL-LR algorithm is comparable to that of the other unrolled networks (MoDL and ADMM-Net), which is superior to that of UNET and SENSE. We note that the performance of all Lipschitz constrained methods degrade gracefully in the presence of Gaussian noise. The experiments show that MOL-LR, which uses the Lipschitz regularization, is less sensitive to adversarial noise than are MoDL and ADMM-Net. We note that the robustness of the original MoDL and ADMM-Net implementations can be improved by Lipschitz regularization (MoDL-LR and ADMM-Net-LR), albeit with a decrease in performance in the noiseless setting. By contrast, the larger number of iterations in MOL translates to improved performance in the noiseless condition, while being robust to input perturbations.}
\label{fig:brain_rob_compare}
\end{figure*}

The comparison of performance of the algorithms on four-fold accelerated knee data is shown in Fig. \ref{fig:knee_perf_compare} and Table \ref{tab:perf_comp_tab}, respectively. Cartesian 1D non-uniform variable density mask has been used for undersampling the data. Table \ref{tab:perf_comp_tab} reports the quantitative performance in terms of mean PSNR and SSIM on 20 subjects. We observe that the performance of MOL-LR is only marginally lower than ten-iteration MoDL and ADMM-Net. The marginally lower performance of MOL-LR can be attributed to the stricter Lipschitz constraint on the CNN blocks, compared to MoDL and ADMM-Net. This is also confirmed by our experiments on Calgary brain data in Fig. \ref{fig:brain_rob_compare}, where the performance of MoDL and ADMM-Net decreases even more with the addition of the Lipschitz constraint (see MoDL-LR and ADMM-Net-LR reconstructions with no input noise). We also note that the reduction in performance is higher for the MOL-SN and DE-GRAD, both of which use spectral normalization. The performance of the UNET is lower than that of the unrolled algorithms MoDL, ADMM-Net, and MOL-LR. The SENSE reconstruction performance is the lowest among all. The comparison of the average runtimes of the algorithms show that the MOL-LR scheme with around 25 DEQ iterations is around 2.5 times higher than MoDL and ADMM-Net with 10 unrolling steps. The qualitative comparisons are shown in Fig. \ref{fig:knee_perf_compare}. The error images show higher errors for MOL-SN, DE-GRAD, SENSE and UNET, while the error images from MOL-LR, MoDL, and ADMM-Net are comparable. We performed statistical tests to compare MOL-LR against the other methods in terms of PSNR and SSIM reported in Table \ref{tab:perf_comp_tab}. This was done for PSNR using linear mixed model analysis with Dunnett’s test for pairwise comparison of means. For SSIM values which did not meet normality assumptions, Friedman’s test was used, with pairwise comparisons tested using Wilcoxon signed-rank test with Bonferroni adjustment applied to the p-values to account for multiplicity. The PSNR comparisons showed statistically significant differences between MOL-LR and each of the other methods. MoDL and ADMM-Net performed only slightly better than MOL-LR, with difference in mean PSNR of +0.40 (95\% confidence interval (CI): 0.16, 0.64; p $<$ 0.0001) and +0.29 (95\% CI: 0.05, 0.53; p $<$ 0.009), respectively. All the other methods underperformed compared to MOL-LR with a much larger mean difference in PSNR, from -1.81 (95\% CI: -1.05, -2.05) for MoDL-LR to -5.30 (95\% CI: -5.03, -5.54) for SENSE. Comparison of SSIM showed no significant difference between MOL-LR and MoDL (median difference 0.000; 95\% CI: -0.001, +0.001; p = 1.00) and ADMM-Net (median difference 0.000; 95\% CI: 0.000, +0.001; p = 1.00). All the other methods had significantly smaller SSIM compared to MOL-LR (p $<$ 0.0001), from -0.003 (95\% CI: -0.002, -0.004) for MoDL-LR and ADMM-Net-LR to -0.007 (95\% CI: -0.006, -0.008) for SENSE and UNET-LR.

We also perform experiments to compare the proposed MOL-LR against unrolled method MoDL and the CS approach SENSE in uniform undersampling setting. The multi-channel knee data is four-fold undersampled using a uniform mask as shown in Fig. \ref{fig:knee_uniform}. Table \ref{tab:perf_comp_uniform_tab} reports the quantitative results in terms of PSNR and SSIM on 20 subjects. The quality of MOL-LR is slightly lower than MoDL with no regularization, which is consistent. Both the methods significantly outperform SENSE which has visible aliasing in the reconstructed image as evident from Fig. \ref{fig:knee_uniform}. It is found through statistical analysis that MoDL is only slightly better than MOL-LR in PSNR (mean difference of +0.59; 95\% CI: 0.43, 0.74; p $<$ 0.0001), but with no significant difference in SSIM (median difference of 0.000; p = 1.00). SENSE underperformed compared to MOL-LR with a mean difference of -5.27 (95\% CI: -5.11, -5.43; p $<$ 0.0001) for PSNR, and median difference of -0.007 (95\% CI: -0.006, -0.008; p $<$ 0.0001) for SSIM (See Table \ref{tab:perf_comp_uniform_tab}). The statistical methods used in this case are the same as the ones used for non-uniform undersampling case, mentioned above.

Quantitative comparisons of the methods on four-fold accelerated Calagary brain data is reported by Table I in the supplementary material. For this case, cartesian 2D non-uniform variable density mask has been used. A similar trend is observed here in terms of performance metrics PSNR, SSIM and run-time of the algorithms. The reconstructed brain images for qualitative comparisons are shown in Fig. \ref{fig:brain_rob_compare} in the main paper and also in Fig. 1 from the supplementary material. The first column of Fig. \ref{fig:brain_rob_compare} shows reconstructions from different methods when no noise has been added. MOL-LR performs comparable to the unrolled algorithms MoDL and ADMM-Net. The Lipschitz constrained MoDL (MoDL-LR) and ADMM-Net (ADMM-Net-LR) show relatively lower performance due to constrained CNNs. Both UNET and UNET-LR show reduction in performance as compared to the unrolled algorithms while MOL-SN and DE-GRAD also show reduced performance due to Lipschitz constraint enforced through spectral normalization. Similar trends are also seen in a different slice in Fig. 1 of the supplementary material.

\subsection{Robustness to input perturbations}
\begin{figure*}[h!]
	\centering
	\includegraphics[width=0.8\textwidth,keepaspectratio=true,trim={1.7cm 11.8cm 10.9cm 12.1cm},clip]{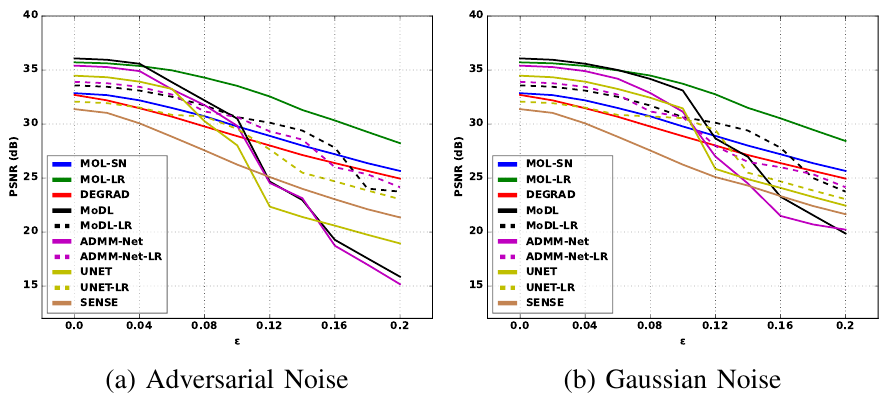}
	\caption{Quantitative comparison of the robustness of different algorithms to worst-case and Gaussian input perturbations. (a) shows the plot between PSNR and $\epsilon$, which is the ratio of the norm of perturbations and the norm of the measurements, and (b) shows a similar plot for Gaussian noise. In the case of Gaussian noise in (b), all the models show an almost linear decrease in performance. By contrast, we note that the PSNR of the models trained without any Lipschitz constraints (solid curves that denote MoDL, ADMM-Net, UNET) drop significantly at around $\epsilon=$10\%. The performance of the Lipschitz-constrained algorithms exhibit linear decay in (a), similar to the Gaussian setting in (b). However, the Lipschitz-constrained versions are associated with a decrease in performance in the noiseless setting, resulting from the constrained CNN. The proposed MOL-LR curves in the Gaussian and worst-case setting are roughly similar, while offering performance similar to the unrolled methods without Lipschitz constraint.}
	\label{fig:brain_robustness_plots}
\end{figure*}

We compare the robustness of the networks on four-fold accelerated brain data in Fig. \ref{fig:brain_rob_compare} and Fig. \ref{fig:brain_robustness_plots} respectively.
Specifically, we study the change in output with respect to the perturbations to input to determine the stability of the models.

The first column in Fig. \ref{fig:brain_rob_compare} shows the reconstructions given by different methods when there is no additional noise in the measured k-space data (no-noise). The second and third column shows reconstructions when the measurements are corrupted by worst-case perturbations (adversarial) with energy as 10\% and 15\% of the energy of the measurements respectively. The fourth column shows reconstructed images when the measurements are corrupted by Gaussian noise with energy as 15\% of the energy of the measurements. Here, MoDL, ADMM-Net and UNET are traditional methods with no Lipschitz constraint. By contrast, MOL-SN and DE-GRAD use spectral normalization, while MOL-LR uses the proposed Lipschitz constraint. MoDL-LR, ADMM-Net-LR, and UNET-LR correspond to the above traditional methods with the proposed Lipschitz constraints added to the CNN blocks. 

 The performance of MOL-LR is comparable to that of MoDL and ADMM-Net in the setting with no additional noise, which is also consistent with the findings in Table \ref{tab:perf_comp_tab} and the PSNR plots in Fig. \ref{fig:brain_robustness_plots}. The improved performance of these unrolled methods over UNET is well established. The MOL-SN and DE-GRAD schemes that use spectral normalization are associated with lower performance. We notice from the last column that the performance of all the Lipschitz-constrained methods only decrease by around $4-5$ dB with the addition of 15\% Gaussian noise. However, the performance of ADMM-Net, MoDL, and UNET drops by around 10 dB with adversarial perturbations of the same amount of norm. By contrast, the performance drop of the Lipschitz-constrained models are largely consistent between the Gaussian and the worst-case setting, indicating that the proposed constraint can stabilize unrolled methods as well. However, we note that ADMM-Net-LR, MoDL-LR, and UNET-LR are associated with a decreased performance in the case with no additional noise. The decrease in performance can be attributed to the more constrained CNN block. The MOL-LR scheme offers performance comparable to the competing methods in the case without additional noise, while it is also more robust to worst-case perturbations. The better performance of MOL-LR compared to other LR methods in the absence of additional noise perturbations can be attributed to the higher number of iterations (nFE=28) compared to the 10 unrolling steps used in those methods. These trends can also be appreciated from the plot of the PSNRs in Fig. \ref{fig:brain_robustness_plots}. (a). The models without Lipschitz constraints (MoDL, ADMM-Net, UNET) exhibit a drastic drop with $\epsilon > 0.1$, whereas SENSE, MOL-LR, and MOL-SN have an approximately linear drop. We note that our theory predicts a linear drop in performance with MOL methods.
MOL-SN and DE-GRAD (blue \& red curves) are the flattest, which can be explained with the smaller Lipschitz constant. Although quantitative analysis of the reconstructions clearly show the proposed MOL-LR to be superior, a rigorous qualitative analysis by radiologists is needed to determine if the MOL-LR reconstructions are fit for clinical purposes.   


\begin{figure*}[h!]%
    \centering
    \subfloat[\centering 4x]{{\includegraphics[scale=0.465,keepaspectratio=true,trim={1.5cm 6.1cm 1.6cm 6.1cm},clip]{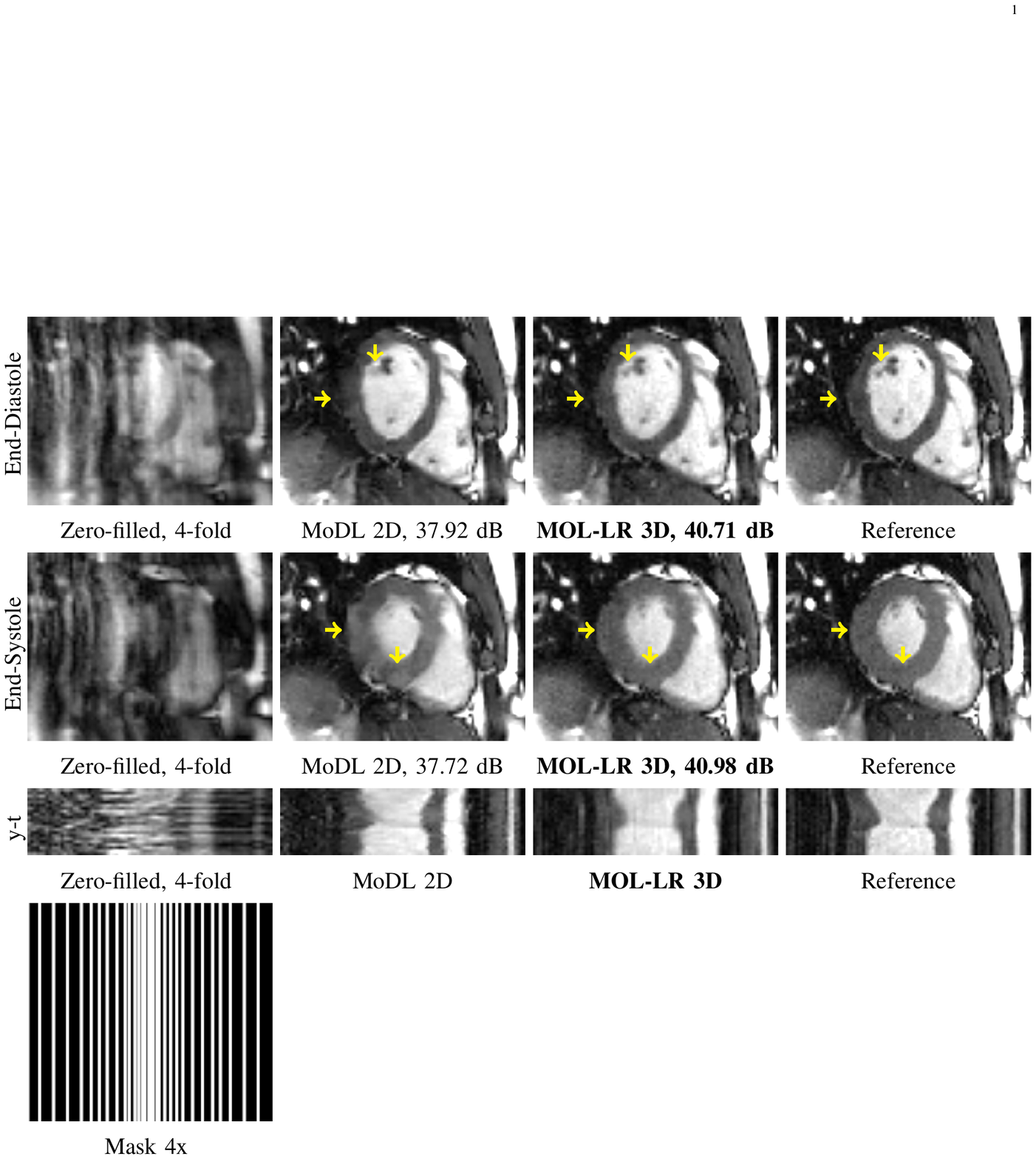}}}%
    \qquad
    \subfloat[\centering 6x]{{\includegraphics[scale=0.465,keepaspectratio=true,trim={1.5cm 6.1cm 1.6cm 6.1cm},clip]{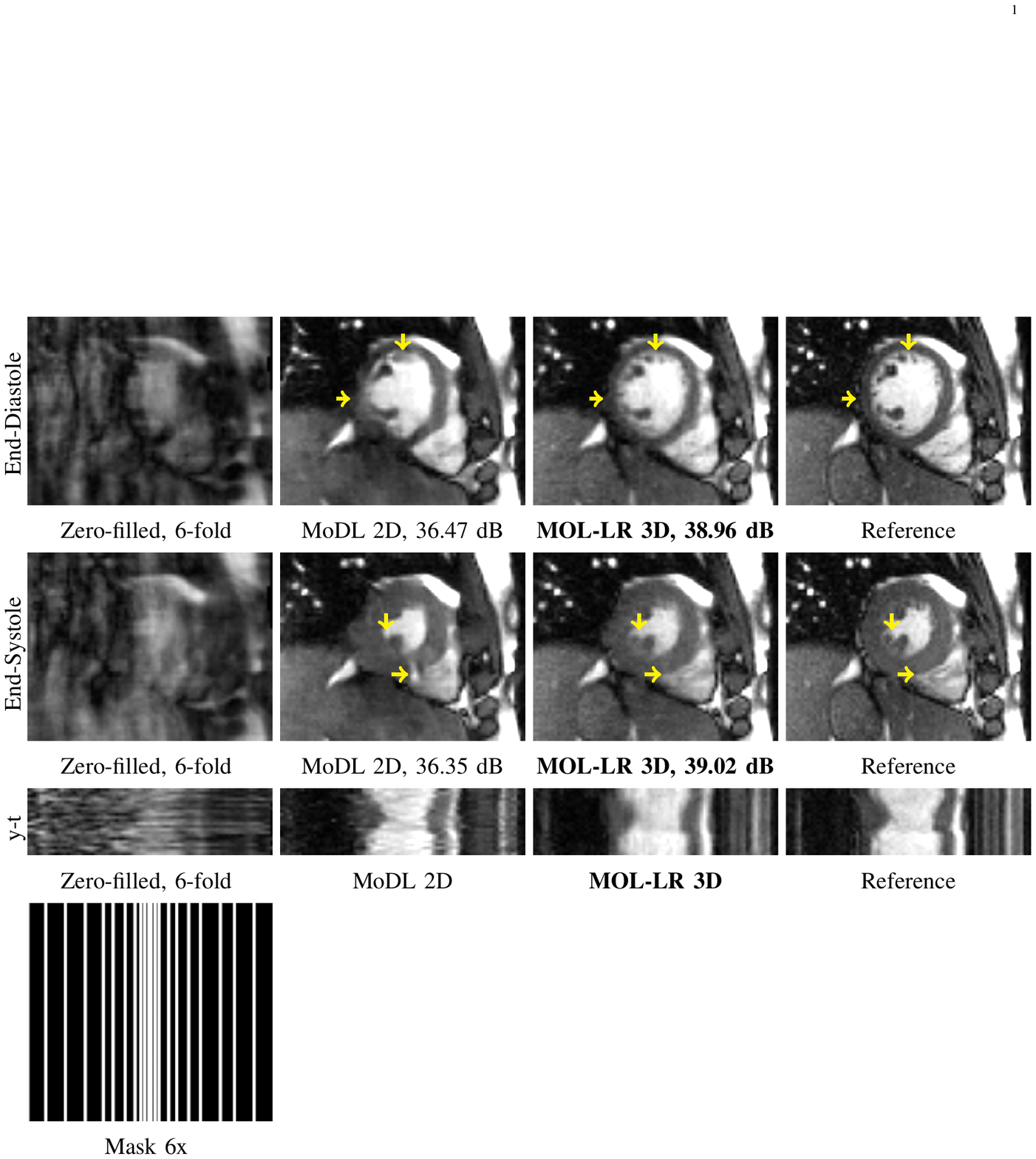}}}%
    \caption{MOL recovery of 2D+time cine data at 4x and 6x accelerations. The PSNR (dB) values are reported for each case. The data is retrospectively undersampled using a Poisson density sampling pattern. For both figures (a) and (b), the top and middle rows correspond to the diastole and systole phases, respectively. For MoDL 2D and MOL-LR 3D, reconstructions (magnitude images) are shown. MOL-LR 3D reconstructs 2D+time data using 3D CNN while MoDL 2D processes each of the temproral frame independently using 2D CNN. Thus, MoDL 2D does not exploit inter-frame depedencies. MoDL 2D has more errors in the boundary of the myocardium than does MOL-LR 3D, as indicated by the yellow arrows. MoDL 2D reconstructs images in the spatial domain only, whereas MOL-LR 3D exploits redundancies in the additional temporal domain, leading to lower errors. The short axis cut looks sharper for MOL-LR 3D and shows improved preservation of wall details.}%
    \label{fig:cardiac}%
\end{figure*}

\begin{table}[h!]
\fontsize{7}{10}
\selectfont
\centering
\begin{tabular}{|c|cc|cc|}
\hline
 \multicolumn{5}{|c|}{2D vs 2D+time Cardiac CINE MRI Recovery} \\ \hline
Acceleration & \multicolumn{2}{c|}{4x} & \multicolumn{2}{c|}{6x} \\  
Methods & PSNR & SSIM & PSNR & SSIM \\ \hline 
Zero-filled & 25.25 & 0.764 & 23.64 & 0.712 \\
MoDL 2D & 38.21 & 0.975 & 36.94 & 0.959  \\
MOL-LR 2D+t & \textbf{40.68} & \textbf{0.988} & \textbf{39.16} & \textbf{0.976}  \\ \hline
\end{tabular}
\vspace{1em}
\caption{Quantitative comparisons on 2D+time CINE data. }
\label{tab:cardiac_comp_tab} 
\end{table}

\subsection{Illustration in high-dimensional settings}

The proposed MOL training approach only requires one physical layer to evaluate the forward and backward propagation; the memory demand of MOL-LR is ten times smaller than that of the unrolled networks with ten iterations. Our 2D experiments show that MOL-LR achieves performance similar to unrolled algorithms with a much lower memory footprint, which makes it an attractive option for large-scale problems. In this section, we illustrate the preliminary applicability of the MOL framework to large-scale problems. The joint recovery of 2D+time data using different undersampling patters for each time point can capitalize on the strong redundancy between the time-frames. However, it is challenging to use conventional unrolled optimization algorithms because of the high memory demand. We compare the performance of a 2D+time version of MOL-LR against a 2D MoDL (ten iterations) for recovery of time series of cardiac CINE MRI.

The reconstruction results for six-fold and four-fold accelerated CINE MRI recovery are shown in Fig. \ref{fig:cardiac}.(a) and Fig. \ref{fig:cardiac}.(b), respectively. The top two rows correspond to the diastole and systole phases, respectively. The third row corresponds to the time series. It is observed from the top two rows that the 2D+time MOL-LR  approach is able to minimize spatial blurring when compared to the 2D MoDL approach. The frame-to-frame changes in aliasing artifacts can also be appreciated from the time plots. Table \ref{tab:cardiac_comp_tab} displays the mean PSNR and SSIM over eight test subjects. MOL-LR 3D outperforms MoDL 2D in terms of PSNR (by $\approx$ 2 dB) and SSIM at both the acceleration factors. The masks are shown in the bottom row. The four-fold and six-fold Poisson density sampling masks used in these experiments have eight lines in the center. While the preliminary experiments in this context are encouraging, more experiments are needed to compare the MOL in this setting to state-of-the-art dynamic MRI methods. We plan to pursue this in the future.

\section{Conclusion}
We introduced a deep monotone operator learning framework for model-based deep learning to solve inverse problems in imaging. The proposed approach learns a monotone CNN in a deep equilibrium algorithm. The DEQ formulation enables forward and backward propagation using a single physical layer, thus significantly reducing the memory demand. The monotone constraint on the CNN allows us to introduce guarantees on the uniqueness of the solution, rapid convergence, and stability of the solution to input perturbations. We introduced two implementations that differ in the way the monotone constraint is imposed. The first approach relies on an exact spectral normalization strategy, while the second method relies on an approximate regularization approach. Our experiments show that both approaches result in convergent algorithms that are more robust to input perturbations than other deep learning approaches. However, the less conservative regularization-based implementation offers improved performance compared to the more constrained spectral normalization approach. The validations in the context of parallel MRI show that the proposed MOL framework provides performance similar to the unrolled MoDL algorithms, but with significantly reduced memory demand and improved robustness to worst-case input perturbations. The memory efficiency of the proposed scheme enabled us to demonstrate the preliminary utility of this scheme in a larger-scale (2D+time) problem. The preliminary comparisons in the super-resolution setting also show that the proposed method is broadly applicable to other linear inverse problems.

\section{Appendix}



\subsection{Proof of Proposition \ref{propv4}}

\begin{proof}
Assume that there exist two fixed points $\mathbf x\neq \mathbf y$ for a specific $\mathbf b$:
\begin{eqnarray}
\lambda \mathbf A^H(\mathbf A\mathbf x - \mathbf b) + \mathcal F(\mathbf x) &=& 0 \\
\lambda \mathbf A^H(\mathbf A\mathbf y - \mathbf b) + \mathcal F(\mathbf y) &=& 0
\end{eqnarray}
which gives
\begin{eqnarray}
\label{zeq}
\mathbf z = \lambda \mathbf A^H(\mathbf A(\mathbf x - \mathbf y)) + \mathcal F(\mathbf x) - \mathcal F(\mathbf y) = \mathbf 0.
\end{eqnarray}

Setting, $\mathbf v= \mathbf x-\mathbf y$, we consider
\begin{eqnarray*}
\Re\left(\langle \mathbf z, \mathbf v  \rangle\right) &=& \underbrace{\Re\left(\Big \langle \lambda \mathbf A^H \mathbf A \mathbf v, \mathbf v \Big \rangle\right)}_{\geq \lambda \mu_{\rm min} \|\mathbf v\|_2^2} +\\&&\qquad \underbrace{\Re\left(\Big \langle \mathcal F(\mathbf x) - \mathcal F(\mathbf y), \mathbf v \Big \rangle\right)}_{\geq m\|\mathbf v\|_2^2} \\
\nonumber
&\geq& (\lambda \mu_{\rm min} + m)~\|\mathbf v\|_2^2  
\end{eqnarray*}
where $\mu_{\rm min}\geq 0$ is the minimum eigenvalue of $\mathbf A^H \mathbf A$ operator and $\mathcal F$ is $m$-monotone. The above relation is true only if $\mathbf v = 0$ or $\mathbf z \neq 0$. The first condition is true if $\mathbf x=\mathbf y$, while the second condition implies that \eqref{zeq} is not true for $\mathbf v \neq 0$ or $\mathbf x \neq \mathbf y$.

We will now present a counter-example to show that the constraint is necessary. Suppose $\mathcal F$ is a linear non-monotone operator, denoted by a symmetric matrix $\mathbf F$ that has a null-space $\mathcal N(\mathbf F)$ which overlaps with null-space $\mathcal N(\mathbf A)$ of $\mathbf A$. Since $\mathbf F$ is not monotone, it will not satisfy $\Re\left(\langle \mathbf F\mathbf v, \mathbf v\rangle\right) >0$, which implies that $\mathbf F$ is not positive definite. If the null-spaces of $\mathbf A $ and $\mathbf F$ overlap, we can choose a $\mathbf v \in \mathcal N(\mathbf A) \bigcap \mathcal N(\mathbf F) $ such that $\langle \mathbf z, \mathbf v\rangle =0$. This counter-example shows that there exist non-monotone operators such that the fixed points are not unique.   
\end{proof}

\subsection{Proof of Proposition \ref{resnet}}
\begin{proof}
Let the Lipschitz constant of $\mathcal H_{\theta}$ is $1 - m$:
\begin{eqnarray}
\label{eqn:lip_h}
\|\mathcal H_{\theta}(\mathbf x) - \mathcal H_{\theta}(\mathbf y)\|_2 \leq \underbrace{(1-m)}_{\epsilon}\|\mathbf x - \mathbf y\|_2, \hspace{6pt} \epsilon > 0. 
\end{eqnarray}
Using Cauchy Schwartz, \\$-\|\mathbf a\|_2\cdot\|\mathbf b\|_2 \leq \Re\left(\left\langle \mathbf a,\mathbf b \right\rangle\right)$ and \eqref{eqn:lip_h}, we have
\begin{equation}\label{cs}
-(1-m) \|\mathbf x-\mathbf y\|_2^2 \leq \Re\left(\langle \mathcal H_{\theta}(\mathbf x) - \mathcal H_{\theta}(\mathbf y), \mathbf x - \mathbf y \rangle\right) 
\end{equation}

We consider the inner product, 
\begin{eqnarray}\nonumber
s & = & \Re\left(\langle \mathcal F(\mathbf x) - \mathcal F(\mathbf y), \mathbf x - \mathbf y \rangle\right) \\ \nonumber
s & =& \Re\left(\langle (\mathcal I - \mathcal H_{\theta})(\mathbf x) - (\mathcal I - \mathcal H_{\theta})(\mathbf y), \mathbf x - \mathbf y \rangle \right)\\\nonumber
& =& \|\mathbf x - \mathbf y \|_2^2 - \Re\left(\Big \langle \mathcal H_{\theta}(\mathbf x) - \mathcal H_{\theta}(\mathbf y), \mathbf x - \mathbf y \Big \rangle \right). \\\label{start}
&\geq& m \|\mathbf x - \mathbf y \|_2^2 
\end{eqnarray}

In the second step, we used \eqref{cs}. The relation \eqref{start} shows that $F$ is $m$-monotone.
The second relation specified by \eqref{lip-relation} can be derived using the triangle equality,  
\begin{eqnarray*}
\| \mathcal F(\mathbf x) - \mathcal F(\mathbf y)\|_2 &=& \| \mathbf x-\mathcal H_{\theta}(\mathbf x) - \mathbf y + \mathcal H_{\theta}(\mathbf y)\|_2 \\ 
 & \leq & \| \mathbf x-\mathbf y\|_2 + \| \mathcal H_{\theta}(\mathbf x) - \mathcal H_{\theta}(\mathbf y)\|_2 \\ 
 & \leq & (2-m) \| \mathbf x-\mathbf y\|_2.
\end{eqnarray*}

\end{proof}

\subsection{Proof of Lemma \ref{lem1}}

\begin{proof}
	Using $\mathcal R = (\mathcal I - \alpha \mathcal F)$, 
	\begin{eqnarray}
            \nonumber
		\|\mathcal R(\mathbf x) - \mathcal R(\mathbf y)\|_2^2
		& = & \|\mathbf x - \mathbf y - \alpha \mathcal F(\mathbf x) + \alpha \mathcal F(\mathbf y)\|_2^2 \\
		\nonumber
		&= & \|\mathbf x - \mathbf y \|_2^2 + \alpha^2 \underbrace{\|\mathcal F(\mathbf x) - \mathcal F(\mathbf y)\|_2^2}_{<(2-m)^2\|\mathbf x-\mathbf y\|_2^2} +\\
            \nonumber
		&&
		\underbrace{- 2\alpha \Re\left(\Big\langle \mathbf x - \mathbf y, \mathcal F(\mathbf x) - \mathcal F(\mathbf y) \Big\rangle\right)}_{<-2\alpha m \|\mathbf x-\mathbf y\|_2^2} \rangle. \\
            && \label{imp} \\
            \nonumber
		&\leq&  \|\mathbf x - \mathbf y \|_2^2 \sqrt{1+\alpha^2 (2-m)^2 - 2\alpha m},
	\end{eqnarray}
	which shows that $L[\mathcal R]=\sqrt{1+\alpha^2 (2-m)^2 - 2\alpha m}$. 
	The first inequality in \eqref{imp} follows from the Lipschitz bound for $\mathcal F$ in (\ref{lip-relation}), while the second one is from the $m$-monotonicity \eqref{eqn:monotonicity_definition} condition on $\mathcal F$.
 
\end{proof}

\subsection{Proof of Proposition \ref{thmiv3}}

\begin{proof}
We first show that the operator $\mathcal T_{\rm MOL}$ in the iterative relation \eqref{fp}:
\begin{equation}
    \mathbf x_{n+1} = \mathcal T_{\rm MOL} (\mathbf x_{n}) + \mathbf z
\end{equation}
is a contraction. In particular, the Lipschitz constant of $\mathcal Q_{\alpha}$ in \eqref{inverse} is given by $L[\mathcal Q_{\alpha}]=\frac{1}{(1+\lambda \mu_{\rm min})}$, where $\mu_{\rm min} \geq 0$ is the minimum eigenvalue of $\mathbf A^H \mathbf A$.

Under the conditions of the theorem, the Lipschitz constant $L[\mathcal I-\alpha \mathcal F]$ is less than one. Combining the two, we have $L[\mathcal T_{\rm MOL}]<1$. If $\mathbf x^*$ is a fixed point, we have $\mathbf x^*=\mathcal T_{\rm MOL}(\mathbf x^*)+\mathbf z$. The result follows by the straightforward application of the Banach fixed-point theorem. 
\end{proof}

\subsection{Proof of Proposition \ref{robustness}}

\begin{proof}
Consider the iterative rule in \eqref{fp},
\begin{eqnarray}\nonumber
\mathbf x_{n} &=& \underbrace{\mathcal Q_{\alpha}(\mathcal I-\alpha \mathcal F)}_{\mathcal T} (\mathbf x_{n-1}) + \alpha \lambda \mathcal Q_{\alpha}(\underbrace{\mathbf A^H \mathbf b}_{\mathbf w})\\\nonumber
&=& \mathcal T^2(\mathbf x_{n-2}) + (\mathcal T+I)~\alpha \lambda  \mathcal Q_{\alpha}(\mathbf w)\\\nonumber
&=& \mathcal T^n(\mathbf x_{0}) + (\mathcal T^{n-1}+ \mathcal T + \ldots +I) ~\alpha \lambda \mathcal Q_{\alpha}(\mathbf w) \\
\label{recursion}
\end{eqnarray}
The Lipschitz bound of $\mathcal Q_{\alpha}$ is, 
\begin{eqnarray}\label{lq}
L_{\mathcal Q_{\alpha}} &=& \frac{1}{1+\lambda \mu_{\rm min}},\\\nonumber
L[\mathcal T] &=& L_{\mathcal Q_{\alpha}}~\sqrt{1 + \alpha^2(2 - m)^2 - 2\alpha m}.\\\label{lt}
\end{eqnarray}
Consider $\mathbf z_1$ and $\mathbf z_2$ as two input measurements with $\boldsymbol\delta =\mathbf z_2-\mathbf z_1$ as the perturbation in the input. Let the corresponding outputs be $\mathbf x_{1,n}$ and $\mathbf x_{2,n}$, respectively, with $\Delta_n= \mathbf x_{2,n}-\mathbf x_{1,n}$ as the perturbation in the output. Thus, the perturbation in the output can be written as
\begin{eqnarray}
\nonumber
\|\Delta_{n}\|_2 &=& \|\mathcal T(\mathbf x_{1,n-1}) - \mathcal T(\mathbf x_{2,n-1}) + \alpha \lambda \mathcal Q_{\alpha}(\boldsymbol \delta)\|_2 \\
\nonumber
 &\leq&  \|\mathcal T(\mathbf x_{1,n-1}) - \mathcal T(\mathbf x_{2,n-1})\|_2 + \alpha \lambda \|\mathcal Q_{\alpha}(\boldsymbol \delta)\|_2 \\
\nonumber
 &\leq&  L[\mathcal T]\|\Delta_{n-1}\|_2 + \alpha \lambda L_{\mathcal Q_{\alpha}}\|\boldsymbol\delta\|_2
\end{eqnarray}
Using \eqref{recursion}, we can expand the above relation as 
\begin{eqnarray*}
\|\Delta_{n}\|_2 &\leq&  (L[\mathcal T])^n\|\Delta_0\|_2 + \\
& &\alpha \lambda ((L[\mathcal T])^{n-1}+(L[\mathcal T])^{n-2}+\ldots 1)L_{\mathcal Q_{\alpha}}\|\boldsymbol \delta\|_2.
\end{eqnarray*}
When $L[\mathcal T] < 1$, the first term vanishes, and we have
\begin{equation*}
\lim_{n\rightarrow \infty}\|\Delta_{n}\|_2 \leq
\frac{\alpha \lambda L_{\mathcal Q_{\alpha}}}{1-L[\mathcal T]}~\|\boldsymbol \delta\|_2.
\end{equation*}
We thus have
\begin{equation}
\lim_{n\rightarrow \infty}\|\Delta_{n}\|_2 = \|\Delta\|_2 \leq
\frac{\alpha \lambda/(1+\lambda \mu_{\rm min}) }{1-\sqrt{1 + \alpha^2(2 - m)^2 - 2\alpha m}}~\|\boldsymbol \delta\|_2.
\end{equation}

\end{proof}
\bibliographystyle{IEEEtran}
\bibliography{ref}

\end{document}



\title{Supplementary material for Memory-efficient model based deep learning with convergence and robustness guarantees}
\author{Aniket Pramanik,~\IEEEmembership{Student Member,~IEEE,} M. Bridget Zimmerman, Mathews Jacob,~\IEEEmembership{Fellow,~IEEE,}
\thanks{Aniket Pramanik and Mathews Jacob are from the Department of Electrical and Computer Engineering at the University of Iowa, Iowa City, IA, 52242, USA (e-mail: aniket-pramanik@uiowa.edu; mathews-jacob@uiowa.edu). M. Bridget Zimmerman is from the Department of Biostatistics at the University of Iowa, Iowa City, IA, 52242, USA (e-mail: bridget-zimmerman@uiowa.edu). This work is supported by grants NIH R01 AG067078 and R01 EB031169.}}

\markboth{Journal of \LaTeX\ Class Files,~Vol.~14, No.~8, August~2021}%
{Shell \MakeLowercase{\textit{et al.}}: A Sample Article Using IEEEtran.cls for IEEE Journals}


\maketitle

\section{DEQ: forward and backward propagation}
This section provides more details on the implementation of the DEQ based MOL algorithm, briefly described in V.A. 
During inference and training, we use the forward iterations $\mathbf x_{n+1} = \mathcal T_{\rm MOL}(\mathbf x_n)+ \mathbf z$ until convergence as shown in the pseudo-code in Algorithm \ref{forward}. The expanded equation is, 
\begin{eqnarray}\nonumber
\label{fp}
\mathbf x_{n+1} & = &  \underbrace{(\mathbf I + \alpha \lambda \mathbf A^H \mathbf A)^{-1}\Big((1-\alpha)\mathbf x_n + \alpha \mathcal H_{\theta}(\mathbf x_n) \Big)}_{\mathcal T_{\rm MOL}(\mathbf x_n)} +\\
&&\qquad \underbrace{(\mathbf I + \alpha \lambda \mathbf A^H \mathbf A)^{-1}\Big( \alpha \lambda \mathbf A^H \mathbf b\Big)}_{\mathbf z}  \\
&=& \mathcal T_{\rm MOL}(\mathbf x_n) + \mathbf z
\end{eqnarray}
from equation (14) in the main paper.
We terminate the algorithm at the $n^{th}$ iteration, when $\mathbf x_n$ satisfies,
	\begin{equation}
		\label{termination}
		e_n = \frac{\|\mathbf x_{n} - \mathbf x_{n-1}\|_2}{\|\mathbf x_{n-1}\|_2} < \kappa .    
	\end{equation} 
We set $\kappa=1\times 10^{-4}$ for the experiments. We denote the fixed point of the algorithm as $\mathbf x^{*}(\mathbf b)$, such that  $\mathbf x^{*}(\mathbf b) \approx \mathcal T_{\rm MOL}\left(\mathbf x^{*}(\mathbf b)\right)+\mathbf z$.

DEQ schemes \cite{gilton2021deep} rely on fixed point iterations for back-propagating the gradients. Using chain rule, the gradient of the loss $\mathcal C$ with respect to the CNN parameters $\theta$ is computed as  $\nabla_{\theta} \mathcal C = ~(\nabla_{\theta} \mathbf x^*)^T  (\nabla_{\mathbf x^*} \mathcal C)$. Using fixed point relation $\mathbf x^* = \mathcal T_{\rm MOL} (\mathbf x^*) + \mathbf z$, one obtains 
$\nabla_{\theta} \mathbf x^* = (\mathcal I - \nabla_{\mathbf x} \mathcal T_{\rm MOL}(\mathbf x)|_{\mathbf x=\mathbf x^*})^{-1} \left(\nabla_{\theta} \mathcal T_{\rm MOL}(\mathbf x^*)\right)$
which translates to 
\begin{equation}
	\label{deq}
	\nabla_{\theta} \mathcal C= \left(\nabla_{\theta} \mathcal T_{\rm MOL}(\mathbf x^*)\right)^\intercal\underbrace{(\mathcal I - \nabla_{\mathbf x} \mathcal T_{\rm MOL}(\mathbf x)|_{\mathbf x=\mathbf x^*})^{-\intercal}   \left(\nabla_{\mathbf x^*} \mathcal C\right)}_{\mathbf q}
\end{equation}
Here, $\mathbf q$ is solved using fixed point iterations \cite{bai2019deep,gilton2021deep} as
\begin{equation}
	\label{deqiter}
	\mathbf q = (\nabla_{\mathbf x} \mathcal T_{\rm MOL}(\mathbf x)|_{\mathbf x=\mathbf x^*})^\intercal ~\mathbf q + \nabla_{\mathbf x^*} \mathcal C
\end{equation}
with initialization as $\mathbf q_0=0$.
This iteration is evaluated until convergence, using the similar termination conditions as in  \eqref{termination}. The computed $\mathbf q$ is substituted in \eqref{deq} to obtain the gradient $	\nabla_{\theta} \mathcal C =  \left(\nabla_{\theta} \mathcal T_{\rm MOL}(\mathbf x^*)\right)^\intercal~\mathbf q$. The pseudo-code for the backpropagation steps are shown in Algorithm \ref{backward}.

\begin{algorithm}

	\caption{: Forward propagation in MOL: input $\mathbf b$}
	\label{forward}
	\begin{algorithmic}[1]
		\State $\mathbf x_0 =  \lambda_0(\mathbf I + \lambda_0 \mathbf A^H \mathbf A)^{-1} \mathbf A^H \mathbf b$
		\State $\mathbf z=(\mathbf I + \alpha \lambda \mathbf A^H \mathbf A)^{-1}\Big( \alpha \lambda \mathbf A^H \mathbf b\Big)$
		\State $e=\infty$
		\While{$e> \kappa$}
		\State $\mathbf x_{\rm old}  = \mathbf x$
		\State $\mathbf x  =\mathcal T_{\rm MOL}(\mathbf x) + \mathbf z$ from \eqref{fp}
		\State $e = \|\mathbf x_{\rm old}-\mathbf x\|_2^2/\|\mathbf x_{\rm old}\|_2^2$ 
		\EndWhile
		\State Return $\mathbf x$
	\end{algorithmic} 
 
\end{algorithm}

\begin{algorithm}

	\caption{: Backpropagation  in MOL: input $\nabla_{\mathbf x^*} \mathcal C$} 
	\label{backward}
	\begin{algorithmic}[1]
		\State $\mathbf q_0 =  0$
		\State $e=\infty$
		\While{$e> \kappa$}
		\State $\mathbf q_{\rm old}  = \mathbf q$
		\State $\mathbf q = (\nabla_{\mathbf x} \mathcal T_{\rm MOL}(\mathbf x)|_{\mathbf x=\mathbf x^*})^\intercal~\mathbf q + \nabla_{\mathbf x^*} \mathcal C$ in \eqref{deqiter}
		\State $e = \|\mathbf q_{\rm old}-\mathbf q\|_2^2/\|\mathbf q_{\rm old}\|_2^2$ 
		\EndWhile
		\State Return $\nabla_{\theta} \mathcal C= \left(\nabla_{\theta} \mathcal T_{\rm MOL}(\mathbf x^*)\right)^\intercal~\mathbf q$
	\end{algorithmic} 

\end{algorithm}

\section{Performance comparison on Parallel MRI from Calgary Brain Data}

\begin{figure*}[t!]
	\centering
	\includegraphics[width=\textwidth,keepaspectratio=true,trim={1.8cm 9cm 0.4cm 8.9cm},clip]{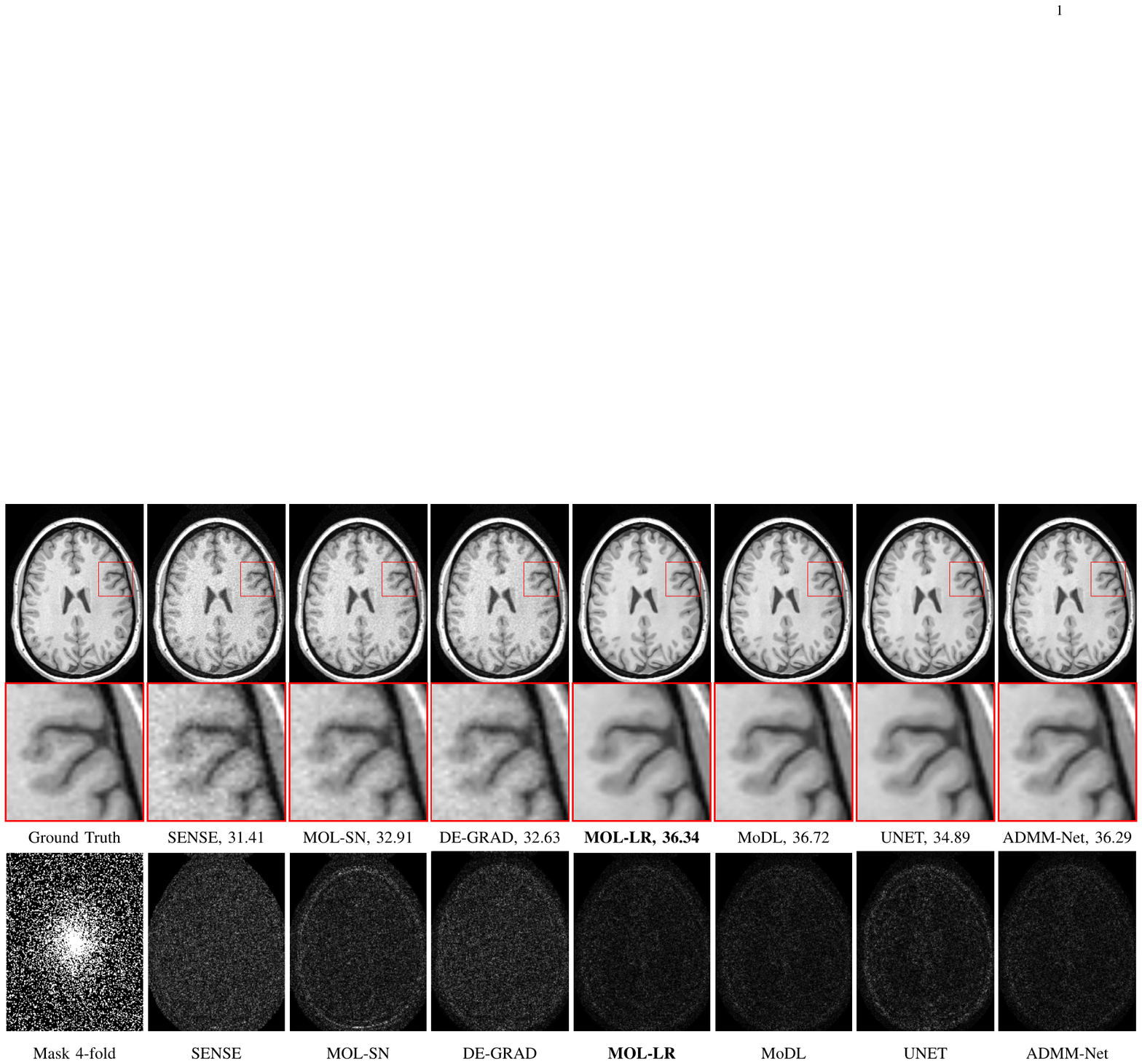}
	\caption{Reconstruction results of 4x accelerated multi-channel brain data. PSNR (dB) values are reported for each case. The image in the first row and column was undersampled using a Cartesian 2D non-uniform variable-density mask as shown in the first column, second row. The top row shows reconstructions (magnitude images), while the bottom row shows corresponding error images. We note that the quality of the \textbf{MOL-LR} reconstructions is comparable to unrolled methods MoDL and ADMM-Net. MOL-SN and DE-GRAD show significantly lower performance due to spectral normalization of weights, resulting in stricter bounds on the Lipschitz constant of its CNN.} 
	\label{fig:brain_perf_compare}
\end{figure*}

Qualitative analysis of the reconstructions from different algorithms on four-fold accelerated brain MRI data is shown in Fig. \ref{fig:brain_perf_compare}. Proposed MOL-LR performs at par with ten-iterations of the unrolled MoDL and ADMM-Net. A slight drop in PSNR for MOL-LR is due to Lipschitz constraint on the CNN block. MOL-SN uses spectral normalization of each layer of the CNN which gives stricter bounds on its Lipschitz and thus provides lower performance. The drop in performance of DE-GRAD is also for similar reasons as in MOL-SN. UNET has a lower performance compared to unrolled algorithms MoDL and ADMM-Net. The error images show high amplitude of errors for SENSE, MOL-SN, DE-GRAD and UNET. Proposed MOL-LR and unrolled MoDL, ADMM-Net show fewer errors in the error images and that too with much lower amplitude. These comparisons show that the Lipschitz regularization strategy offers better performance than spectral normalization. Similar trends are also visible in the quantitative analysis of these methods over 20 subjects as reported in Table \ref{tab:perf_comp_tab}.

\begin{table}[ht!]
\fontsize{7}{10}
\selectfont
\centering
\begin{tabular}{|c|ccc|}
\hline
& \multicolumn{3}{c|}{Four-fold Brain MRI} \\ \hline
Methods & PSNR & SSIM & Run-time \\ \hline 
SENSE & 31.13 $\pm$ 1.67 & 0.982 $\pm$ 0.025 & \textbf{0.06s} \\
MOL-SN & 33.05 $\pm$ 1.13 & 0.986 $\pm$ 0.014 & 0.29s \\
DE-GRAD & 32.84 $\pm$ 1.21 & 0.985 $\pm$ 0.016 & 0.24s \\
MOL-LR & 36.56 $\pm$ 0.82 & 0.992 $\pm$ 0.008 & 0.35s \\ 
MoDL & \textbf{36.89 $\pm$ 0.75} & \textbf{0.993 $\pm$ 0.007} & 0.15s \\
MoDL-LR & 34.63 $\pm$ 1.02 & 0.990 $\pm$ 0.011 & 0.15s \\
ADMM-Net & 36.77 $\pm$ 0.76 & 0.993 $\pm$ 0.008 & 0.16s \\
ADMM-Net-LR & 34.51 $\pm$ 1.03 & 0.990 $\pm$ 0.011 & 0.16s \\
UNET & 34.69 $\pm$ 1.03 & 0.988 $\pm$ 0.011 & 0.07s \\
UNET-LR & 33.42 $\pm$ 1.15 & 0.986 $\pm$ 0.014 & 0.07s \\ \hline
\end{tabular}
\vspace{1em}
\caption{Quatitative comparisons on 2D datasets with 4-fold undersampling using cartesian 2D non-uniform variable density mask. PSNR in dB, SSIM and mean run-time per slice in seconds are reported. The PSNR and SSIM values are in mean $\pm$ standard deviation format.}
\label{tab:perf_comp_tab} 
\end{table}

\section{Illustration in Image Super-resolution Setting}
\begin{figure}[!ht]
	\centering
	\includegraphics[width=0.47\textwidth,keepaspectratio=true,trim={1.8cm 11.7cm 11cm 12cm},clip]{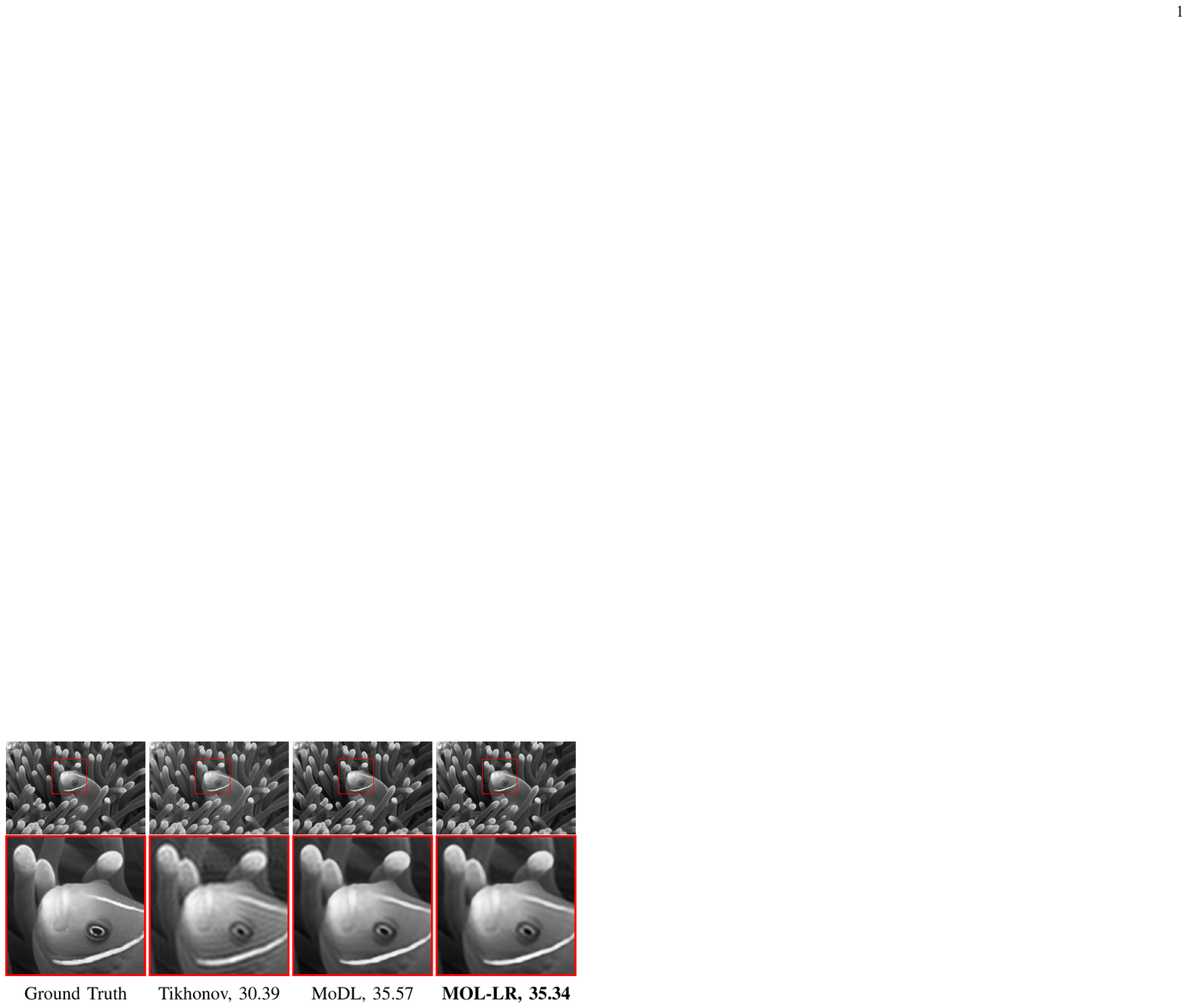}
	\caption{Results of 3x super-resolution reconstruction of natural grayscale images. PSNR (dB) values are reported for each case. The proposed MOL-LR is compared against MoDL and Tikhonov regularized reconstructions. The Tikhonov reconstruction shows blurring and Gibbs ringing, which is reduced significantly in MOL-LR and MoDL, which show similar performance in terms of PSNR.}
	\label{fig:sr}
\end{figure}

The proposed theory and algorithms are broadly applicable to general linear inverse problems. We now show the preliminary feasibility of the proposed MOL-LR approach in three-fold super-resolution image reconstruction in Fig. \ref{fig:sr}. MOL-LR is compared against MoDL and Tikhonov regularized reconstructions, respectively. Similar to the parallel MRI setting, MoDL and MOL-LR offer similar performance while outperforming the Tikhonov regularized reconstruction in terms of PSNR. There is visible aliasing and blurring in Tikhonov, which is reduced to an appreciable extent in MoDL and MOL-LR. All the models have been trained and tested on the Berkeley Segmentation Dataset and Benchmark \cite{MartinFTM01}. The dataset consists of 300 images, out of which 200 are split for training and the remaining ones for testing. While the preliminary experiments in this context are encouraging, more experiments are needed to compare MOL to state-of-the-art super-resolution methods. We plan to pursue this in the future.

\bibliographystyle{IEEEtran}
\bibliography{ref}